\documentclass[manuscript,screen,review=False]{acmart}

\AtBeginDocument{%
  \providecommand\BibTeX{{%
    \normalfont B\kern-0.5em{\scshape i\kern-0.25em b}\kern-0.8em\TeX}}}

\setcopyright{acmcopyright}
\copyrightyear{2021}
\acmYear{2021}
\acmDOI{https://doi.org/}

\begin{document}

\title[GANs and its applications in a wide variety of disciplines - From Medical to Remote Sensing]{A review of Generative Adversarial Networks (GANs) and its applications in a wide variety of disciplines - From Medical to Remote Sensing}

\author{Ankan Dash}
\email{ad892@njit.edu}
\author{Junyi Ye}
\email{jy394@njit.edu}
\author{Guiling Wang}
\email{guiling.wang@njit.edu}

\affiliation{%
  \institution{Department of Computer Science, New Jersey Institute of Technology}
  \streetaddress{323 Dr Martin Luther King Jr Blvd}
  \city{Newark}
  \state{NJ}
  \country{USA}
  \postcode{07102}
}


\begin{abstract}
We look into Generative Adversarial Network (GAN), its prevalent variants and applications in a number of sectors. GANs combine
two neural networks that compete against one another using zero-sum game theory, allowing them to create much crisper and discrete
outputs. GANs can be used to perform image processing, video generation and prediction, among other computer vision applications.
GANs can also be utilised for a variety of science-related activities, including protein engineering, astronomical data processing,
remote sensing image dehazing, and crystal structure synthesis. Other notable fields where GANs have made gains include finance,
marketing, fashion design, sports, and music. Therefore in this article we provide a comprehensive overview of the applications of
GANs in a wide variety of disciplines. We first cover the theory supporting GAN, GAN variants, and the metrics to evaluate GANs.
Then we present how GAN and its variants can be applied in twelve domains, ranging from STEM fields, such as astronomy and
biology, to business fields, such as marketing and finance, and to arts, such as music. As a result, researchers from other fields may
grasp how GANs work and apply them to their own study. To the best of our knowledge, this article provides the most comprehensive
survey of GAN’s applications in different field.
\end{abstract}



\maketitle

\section{Introduction}
Generative Adversarial Networks\cite{GoodfellowNIPS2014} or GANs belong to the family of Generative models\cite{GM}. Generative Models try to learn a probability density function from a training set and then generate new samples that are drawn from the same distribution. GANs generate new synthetic data that resembles real data by pitting two neural networks (the Generator and the Discriminator) against each other. The Generator tries to capture the true data distribution for generating new samples. The Discriminator, on the other hand, is usually a Binary classifier that tries to discern between actual and fake generated samples as precisely as possible.

Over the last few years, GANs have made substantial progress. Due to hardware advances, we can now train deeper and more sophisticated Generator and Discriminator neural network architectures with increased model capacity. GANs have a number of distinct advantages over other types of generative models. Unlike Boltzmann machines\cite{Hinton2010BoltzmannMachines}, GANs do not require Monte Carlo approximations in order to train, and GANs use back-propagation and do not require Markov chains.

GANs have gained a lot of traction in recent years and have been widely employed in a variety of disciplines, with the list of fields in which GANs can be used fast expanding. GANs can be used for data generation and augmentation(\cite{PROGAN:DBLP:journals/corr/abs-1710-10196},\cite{radford2016unsupervised}), image to image translation(\cite{Pix2Pix8100115},\cite{CycleGAN2017}), image super resolution(\cite{ledig2017photorealistic},\cite{TSRGANapp10051729}) to name a few.
It is this versatile nature, that has allowed GANs to be applied in completely non-aligned domains such as medicine and astronomy.

There have been a few surveys and reviews about GANs due to their tremendous popularity and importance. However, the majority of past papers have concentrated on two distinct aspects: first, describing GANs and their growth over time, and second, discussing GANs' use in image processing and computer vision applications(\cite{8833686},\cite{Alqahtani2021},\cite{9182058},\cite{gui2020review},\cite{AGGARWAL2021100004}). As a consequence, the focus has been less on describing GAN applications in a wide range of disciplines.
Therefore, we'll present a comprehensive review of GANs in this first-of-its-kind article. We'll look at GANs and some of the most widely used GAN models and variants, as well as a number of evaluation metrics, GAN applications in a variety of 12 areas (including image and video related tasks,  medical and healthcare, biology, astronomy, remote sensing, material science, finance, marketing, fashion, sports and music), GAN challenges and limitations, and GAN future directions.

\textbf{Some of the major contributions of the paper are highlighted below:}
\begin{itemize}
    \item Describe the wide range of GAN applications in engineering, science, social science, business, art, music and sports. As far as we know, this is the first review paper to cover GAN applications in such diverse domains. This review will assist researchers of various backgrounds in comprehending the operation of GANs and discovering about their wide array of applications.
    \item Evaluation of GANs include both qualitative and quantitative methods. This survey provides a comprehensive presentation of quantitative metrics that are used to evaluate the performance of GANs in both computer vision and time series data analysis. We include evaluation metrics for GANs' application in time series data which are not discussed in other GAN survey paper. To the best of our knowledge, this is the first survey paper to present time series data evaluation metrics for GANs.

\end{itemize}

We have organized the rest of the article as follows: Section 2 presents the basic working of GANs, and the most commonly used GAN variants and their descriptions. Section 3 summarises some of the frequently used GAN evaluation metrics. Section 4 describes the extensive range of applications of GANs in a wide variety of domains. We also provide a table at the end of each subsection summarizing the application area and the corresponding GAN models used. Section 5 discusses some of the difficulties and challenges that are encountered during the training of GANs. Apart from this, we present a short summary concerning the future direction of GAN development. Section 6 provides concluding remarks.

\section{GAN, GAN variants and extensions}
In this section, we describe GANs, the most common GAN models and extensions. Following a description of GAN theory, we go over twelve GAN variants that serve as foundations or building blocks for many other GAN models. There are a lot of articles on GANs, and a lot of them have named-GANs, which are models that have a specific name that usually contains the word ``GAN''. We've focused on twelve specific GAN variants. The reader will obtain a better knowledge of the core aspects of GANs by reading through these twelve GAN variants, which will help them navigate other GAN models.

\subsection{GAN basics}
Generative Adversarial Networks were developed by Ian Goodfellow et al.\cite{GoodfellowNIPS2014} in the year 2014. GANs belong to the class of Generative models\cite{GM}.
GANs are based on the min-max, zero-sum game theory. For this, GANs consist of two neural networks: one is the Generator and the other is the Discriminator. The goal of the Generator is to learn to generate fake sample distribution to deceive the Discriminator whereas the goal of the Discriminator is to learn to distinguish between real and fake distribution generated by the Generator.

\subsubsection{Network architecture and learning}
\begin{figure}[h!]
    \centering
    \includegraphics[width=\textwidth]{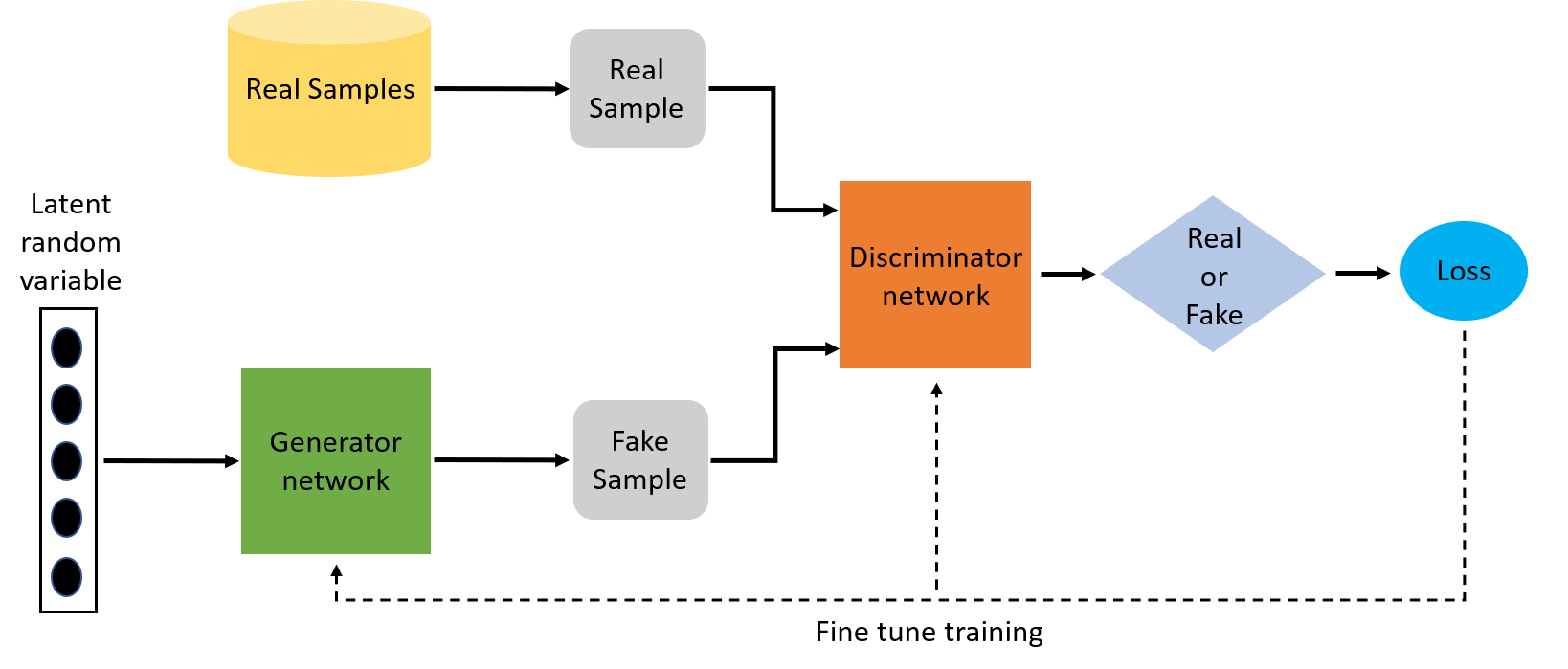}
    \caption{Basic GAN architecture}
    \label{fig:GAN}
\end{figure}

The general architecture of GAN which is comprised of the Generator and the Discriminator is shown in
Figure \ref{fig:GAN}. The Generator (G) takes in as input some random noise vector Z and then tries to generate an image using this noise vector indicated as G(z). The generated image is then passed to the Discriminator and based on the output of the Discriminator the parameters of the Generator are updated. The Discriminator (D) is a binary classifier which simultaneously takes a look at both real and fake samples generated by the Generator and tries to decide which ones are real and which ones are fake. Given a sample image X, the Discriminator models the probability of the image being fake or real. The probabilities are then passed back to the Generator as feedback.

Over time each of the Generator and the Discriminator model tries to one up each other by competing against each other this is where the term ``adversarial'' of Generative Adversarial Networks comes from, and the optimization is based on the minimax game problem. During training both the Generator's and Discriminator's parameters are updated using back propagation with the ultimate goal of the Generator is to be able to generate realistic looking images and the Discriminator to get progressively better at detecting generated fake images from real ones.

GANs use the Minimax loss function which was introduced by Goodfellow et al. when they introduced GANs for the first time. The Generator tries to minimize the following function while the Discriminator tries to maximize it. The Minimax loss is given as,
\begin{equation}
    Min_G Max_D f(D,G) = \mathbb{E}_x[log(D(x))] + \mathbb{E}_z[log(1-D(G(z)))].
\end{equation}
Here, $E_x$ is the expected value over all real data samples, $D(x)$ is the probability estimate of the Discriminator if $x$ is real, $G(z)$ is the output of the Generator for a given random noise vector $z$ as input, $D(G(z))$ is the Discriminator's probability estimate if the fake generated sample is real, $E_z$ is the expected value over all random inputs to the Generator.

\subsection{Conditional  Generative  Adversarial  Nets (cGAN)}
Conditional Generative Adversarial Nets\cite{cGANmirza2014conditional} or cGANs are an extension of GANs for conditional sample generation. This gives control over the modes of data being generated. cGANs use some extra information $y$, such as class labels or other modalities, to perform conditioning by concatenating this extra information $y$ with the input and feeding it into both the Generator G and the Discriminator D.
The Minimax objective function can be modified as shown below,

\begin{equation}
    Min_G Max_D f(D,G) = \mathbb{E}_x[log(D(x|y))] + \mathbb{E}_z[log(1-D(G(z|y)))]
\end{equation}

\begin{figure}[h!]
    \centering
    \includegraphics[width=10cm, height=9cm]{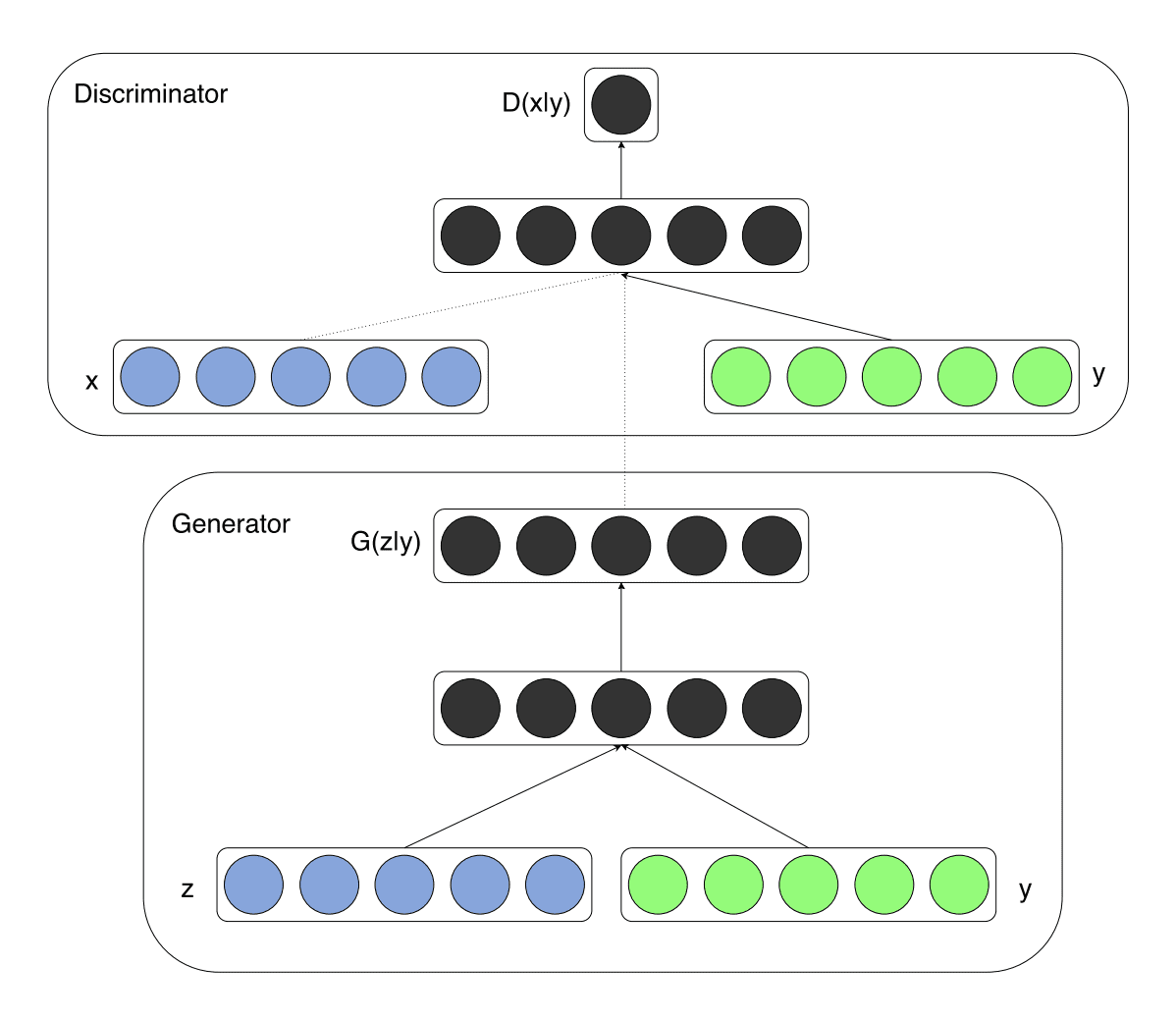}
    \caption{cGAN architecture\cite{cGANmirza2014conditional}}
    \label{fig:cGAN}
\end{figure}

\subsection{Wasserstein GAN (WGAN)}
The authors of WGAN\cite{WGAN_pmlr-v70-arjovsky17a} introduced a new algorithm which gave an alternative to traditional GAN training. They showed that their new algorithm improved the stability of model learning and prevent problems such as mode collapse. For the critique model, WGAN uses weight clipping, which ensures that weight values (model parameters) stay within pre-defined ranges. The authors found that Jensen-Shannon divergence is not ideal for measuring the distance of the distribution of the disjoint parts. Therefore they used the Wasserstein distance which uses the concept of Earth mover's(EM) distance instead to measure the distance between the generated and the real data distribution and while training the model tries to maintain One-Lipschitz continuity\cite{NIPS2017_892c3b1c}.

The EM or Wasserstein distance for the real data distribution $P_r$ and the generated data distribution $P_g$ is given as
\begin{equation}
    W(P_r,P_g) = inf_{\gamma\varepsilon\Pi(P_r,P_g)}\mathbb{E}_{(x,y)\sim r}[\|x-y\|]
\end{equation}
where $\Pi(P_r,P_g)$ denotes the set of all joint distributions $\gamma(x, y)$ whose marginals are respectively $P_r$ and $P_g$. However, the equation for the Wasserstein distance is highly intractable. Therefore the authors used the Kantorovich-Rubinstein duality to approximate the Wasserstein distance as 
\begin{equation}
    max_{w\varepsilon\omega}\mathbb{E}_{x\sim P_r}[f_w(x)]-\mathbb{E}_{z\sim p(z)}[f_w(G(z))]
\end{equation}

where $(f_w)_{w\varepsilon\omega}$ represents a parameterized family of functions that are all K-Lipschitz for some K. The Discriminator D's goal is to optimize this parameterized function which represents the approximated Wasserstein distance. The goal of the Generator G is to miminize the above Wasserstein distance equation such that the generated data distribution is as close as possible to the real distribution. The overall WGAN objective function is given as
\begin{equation}
    min_G max_{w\varepsilon\omega}\mathbb{E}_{x\sim P_r}[f_w(x)]-\mathbb{E}_{z\sim p(z)}[f_w(G(z))]
\end{equation}
or 
\begin{equation}
    min_G max_D\mathbb{E}_{x\sim P_r}[f_w(x)]-\mathbb{E}_{z\sim p(z)}[f_w(G(z))]
\end{equation}
Even though WGAN improved training stability and alleviated problems such as mode collapse however enforcing the Lipschitz constraint is a challenging task. WGAN-GP\cite{NIPS2017_892c3b1c} proposes an alternative to clipping weights by using gradient penalty to penalize the norm of gradient of the critic with respect to its input.

\subsection{Unsupervised Representation Learning with Deep Convolutional Generative Adversarial Networks (DCGANs)}
\begin{figure}[h!]
    \centering
    \includegraphics[width=12cm, height=5cm]{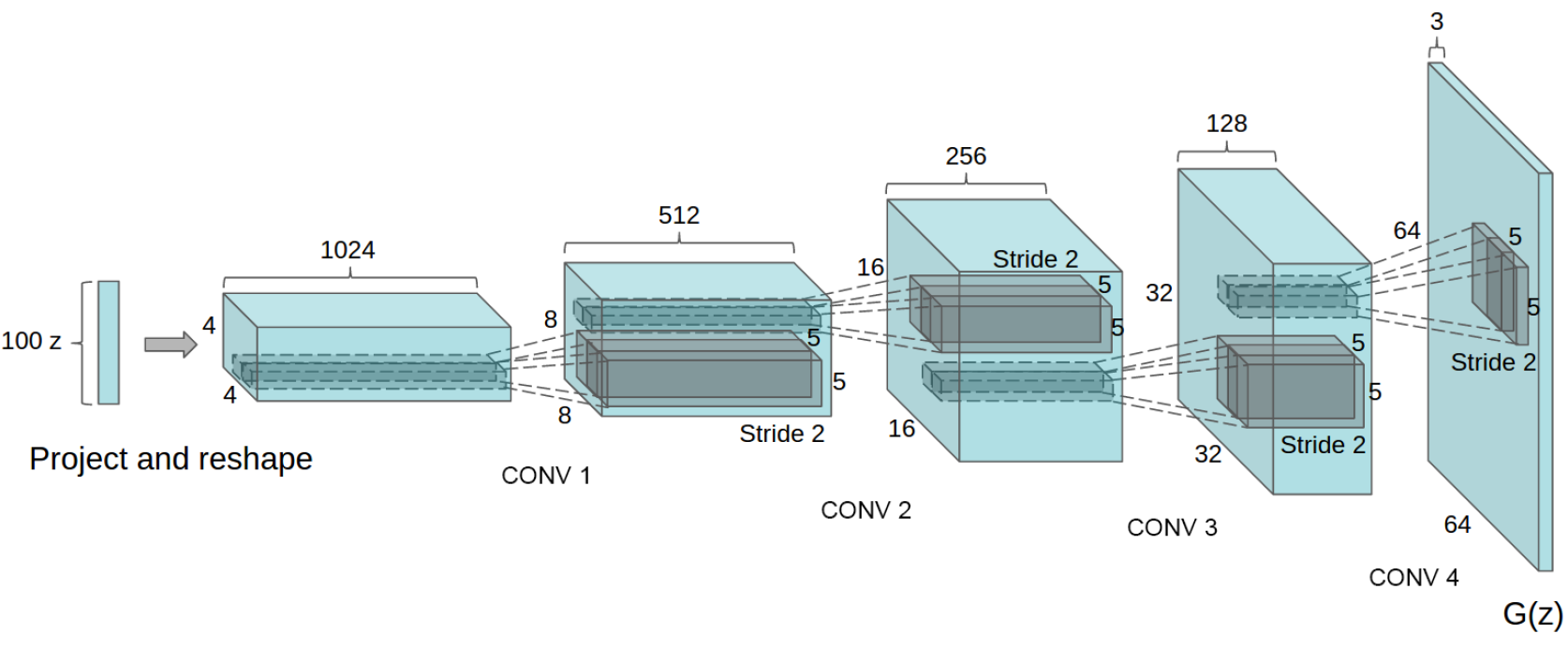}
    \caption{DCGAN Generator architecture\cite{radford2016unsupervised}}
    \label{fig:DCGAN}
\end{figure}

Radford et al.\cite{radford2016unsupervised} introduced the deep convolutional generative adversarial networks or DCGANs. As the name suggests DCGANs use deep convolutional neural networks for both the Generator and Discriminator models. The original GAN architecture used only multi-layer perceptrons or MLPs but since CNNs are better at images than MLP, the authors of DCGAN used CNN in the Generator G and Discriminator D neural network architecture. Three key features of the DCGANs neural network architecture are listed as follows
\begin{itemize}
    \item  First, for the Generator shown in Figure \ref{fig:DCGAN}, convolutions are replaced with transposed convolutions, so the representation at each layer of the Generator is successively larger, as it maps from a low-dimensional latent vector onto a high-dimensional image. Replacing any pooling layers with strided convolutions (Discriminator) and fractional-strided convolutions (Generator).
    \item Second, use batch normalization in both the Generator and the Discriminator.
    \item Third,  use ReLU activation in Generator for all layers except for the output, which uses Tanh. Use LeakyReLU activation in the Discriminator for all layers. 
    \item Fourth, use the Adam optimizer instead of SGD
    with momentum.
\end{itemize}
All of the above modifications rendered DCGAN to achieve stable training. DCGAN was important because the authors demonstrated that by enforcing certain constraints we can develop complex high quality Generators.  The authors also made several other modifications to the vanilla GAN architecture. 

\subsection{Progressive Growing of GANs for Improved Quality, Stability, and Variation (ProGAN)}
\begin{figure}[h!]
    \centering
    \includegraphics[width=12cm, height=6cm]{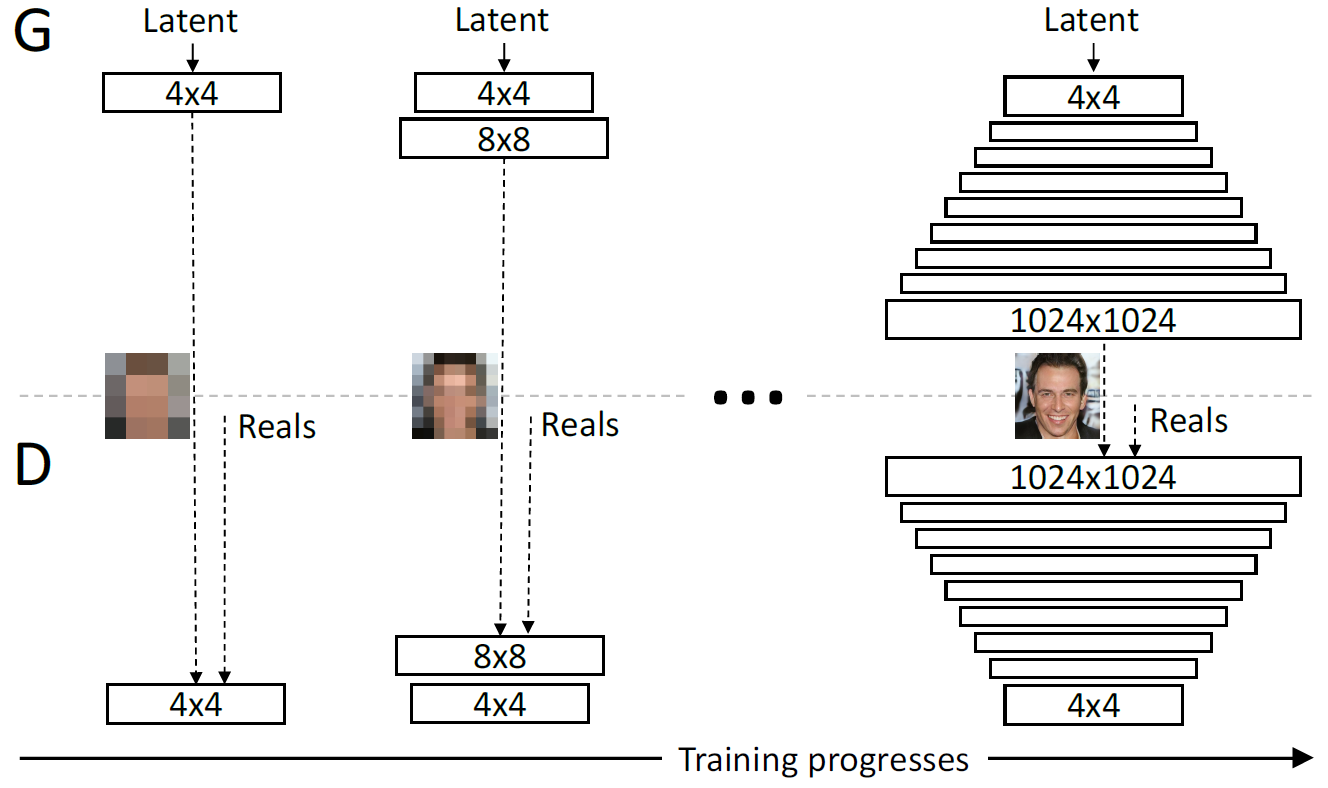}
    \caption{ProGAN architecture\cite{PROGAN:DBLP:journals/corr/abs-1710-10196}}
    \label{fig:ProGAN}
\end{figure}

Karras et al.\cite{PROGAN:DBLP:journals/corr/abs-1710-10196} introduced a new training methodology for training GANs to generate high resolution images. The idea behind ProGAN is to be able to synthesize high resolution and high quality images via the incremental (gradual) growing of the Discriminator and the Generator networks during the training process. ProGAN makes it easier for the Generator to generate higher resolution images by gradually training it from lower resolution images to those higher resolution images (see Figure \ref{fig:ProGAN}.). That is in a progressive GAN, the Generator's first layers produce very low-resolution images, and subsequent layers add details. Training is considerably stabilised by the progressive learning process.

\subsection{Interpretable Representation Learning by Information Maximizing Generative Adversarial Nets (InfoGAN)}
The key motivation behind  InfoGAN\cite{DBLP:journals/corr/ChenDHSSA16} is to enable GANs to learn disentangled representations and have control over the properties or features of the generated images in an unsupervised manner.  To do this instead of using just a noize vector $z$ as input the authors decompose the noise vector into two parts first being the traditional noise vector $z$ and second as new ``latent code vector'' $c$. This code has a predictable effect on the output images. The objective function for InfoGAN\cite{DBLP:journals/corr/ChenDHSSA16} is given as, 
\begin{equation}
    Min_G Max_D f_1(D,G) = f(D,G) - \lambda I(c;G(z,c))
\end{equation}

where $\lambda$ is the regularization parameter, $I(c;G(z,c))$ is the mutual information between the latent code $c$ and the Generator output $G(z,c)$. The idea is to maximize the mutual information between the latent code and the Generator output. This encourages the latent code $c$ to contain as much as possible, important and relevant features of the real data distributions. However it is not practical to calculate the mutual information $I(c;G(z,c))$ explicitly as it requires the posterior $P(c|x)$, therefore a lower bound for $I(c;G(z,c))$ is approximated. This can be achieved by defining an auxiliary distribution $Q(c|x)$ to approximate $P(c|x)$. Thus the final form of the objective function is then given by this lower-bound approximation to the Mutual Information:

\begin{equation}
    Min_G Max_D f_1(D,G) = f(D,G) - \lambda L_1(c;Q)
\end{equation}
where $L_1(c;Q)$ is the lower bound for $I(c;G(z,c))$.
If we compare the above equation to the original GAN objective function we realize that this framework is implemented by merely adding a regularization term to the original GAN’s objective function.

\subsection{StackGAN: Text to Photo-realistic Image Synthesis with Stacked Generative Adversarial Networks (StackGAN)}
\begin{figure}[h!]
    \centering
    \includegraphics[width=\textwidth]{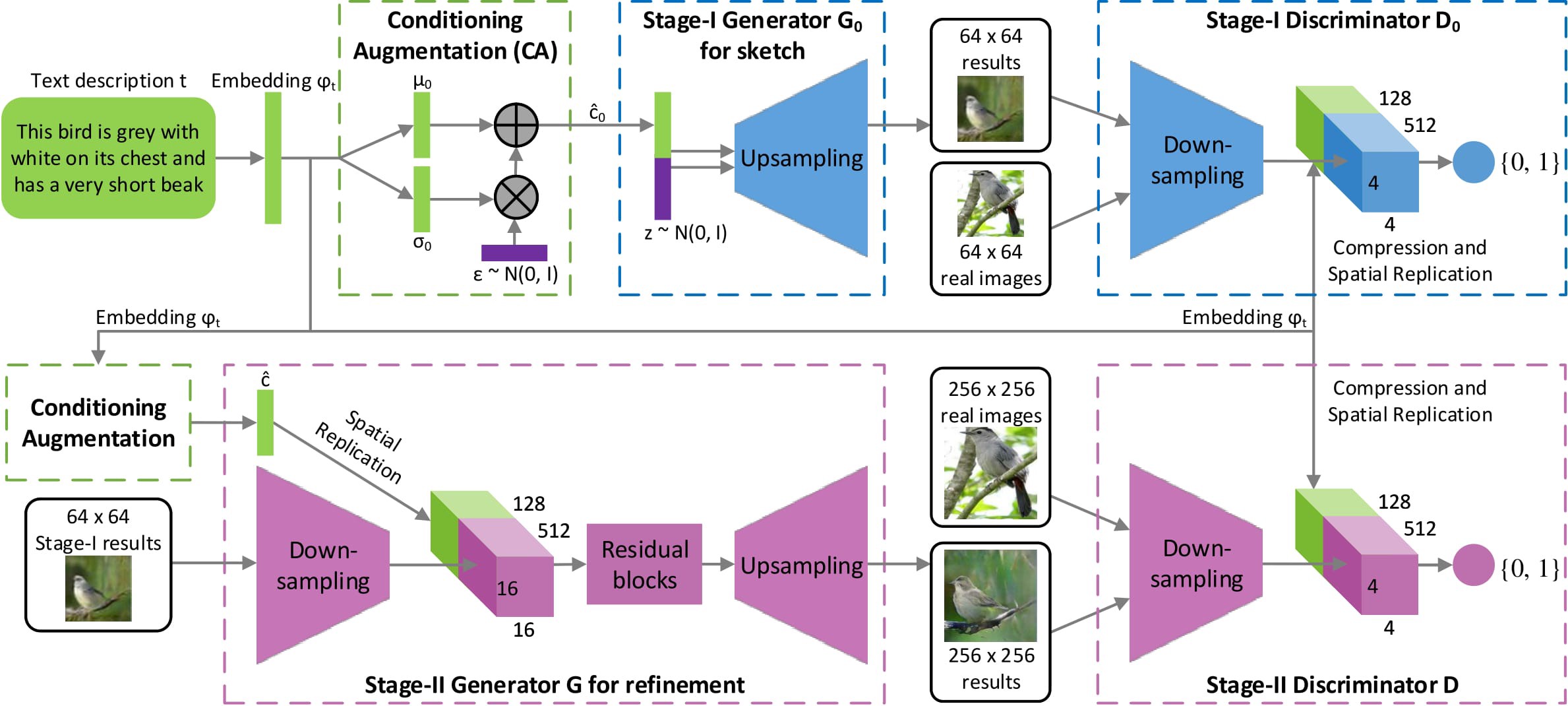}
    \caption{StackGAN architecture\cite{StackGAN8237891}}
    \label{fig:StackGAN}
\end{figure}

StackGAN\cite{StackGAN8237891} shown in Figure~ \ref{fig:StackGAN}, takes in as input a text description and then synthesizes high quality images using the given text description. The authors proposed StackGAN to generate 256×256 photo-realistic images based on text descriptions. To generate photo-realistic images StackGAN uses a sketch-refinement process, StackGAN decomposes the difficult problem into more manageable sub-problems by using Stacked Generative Adversarial Networks. The \textbf{Stage-I GAN} creates Stage-I low-resolution images by sketching the object's primitive shape and colours based on the given text description.
The \textbf{Stage-II GAN} generates high-resolution images with photo-realistic details using Stage-I results and text descriptions as inputs.

To be able to do this, StackGAN architecture consists of the following components: (a) Input variable length text description is converted into a fixed length vector embedding. (b) Conditioning Augmentation. (c) Stage I Generator: Generates (128x128) images (d) Stage I Discriminator (e) Stage II Generator: Generates (256x256) images. (f) Stage II Discriminator

The variable length text description is first converted to a vector embedding which is non-linearly transformed to generate conditioning latent variables as the input of the Generator. Filling the latent space of the embedding with randomly generated fillers is a trick used in the paper to make the data manifold more continuous and thus more conducive to later training. They also add the Kullback-Leibler divergence of the input Gaussian distribution and the  Gaussian distribution as a regularisation term to the Generator's training output, to make the data manifold more continuous and training-friendly.

The Stage I GAN uses the following objective function:
\begin{equation}
    \mathcal{L}_{D_{0}}=\mathbb{E}_{\left(I_{0}, t\right) \sim p_{\text {data }}}\left[\log D_{0}\left(I_{0}, \phi_{t}\right)\right]+\mathbb{E}_{z \sim p_{z}, t \sim p_{\text {data }}}\left[\log \left(1-D_{0}\left(G_{0}\left(z, \hat{c}_{0}\right), \phi_{t}\right)\right)\right]
\end{equation}

\begin{equation}
    \mathcal{L}_{G_{0}}=\mathbb{E}_{z \sim p_{z}, t \sim p_{\text {data }}}\left[\log \left(1-D_{0}\left(G_{9}\left(z, \hat{c}_{0}\right), \phi_{t}\right)\right)\right]+\lambda D_{K L}\left(\mathcal{N}\left(\mu_{0}\left(\phi_{t}\right), \Sigma_{0}\left(\phi_{t}\right)\right) \| \mathcal{N}(0, I)\right)
\end{equation}

The Stage II GAN uses the following objective function:

\begin{equation} 
    \mathcal{L}_{D}=\mathbb{E}_{(I, t) \sim p_{\text {data }}}\left[\log D\left(I, \phi_{t}\right)\right]+\mathbb{E}_{s_{0} \sim p_{G_{0}}, t \sim p_{\text {data }}}\left[\log \left(1-D\left(G\left(s_{0}, \hat{c}\right), \phi_{t}\right)\right)\right]
\end{equation}

\begin{equation}
    \mathcal{L}_{G}=\mathbb{E}_{s_{0} \sim p_{G_{0}}, t \sim p_{\text {data }}}\left[\log \left(1-D\left(G\left(s_{0}, \hat{c}\right), \phi_{t}\right)\right)\right]+\lambda D_{K L}\left(\mathcal{N}\left(\mu\left(\phi_{t}\right), \Sigma\left(\phi_{t}\right)\right) \| \mathcal{N}(0, I)\right)
\end{equation}
where $\phi_t$ is the text embedding of the given description, $p_z$ is Gaussian distribution, $\hat{c}_{0}$ is sampled from a Gaussian distribution from which $\phi_t$ is drawn. $s_{0} = G_{0}(z, \hat{c}_{0})$ and  $\lambda=1$.

\subsection{Image-to-Image Translation with Conditional Adversarial Networks (pix2pix)}
pix2pix\cite{Pix2Pix8100115} is a conditional generative adversarial network(cGAN\cite{cGANmirza2014conditional}) for solving general purpose image-to-image translation problems. 
The GAN consists of a Generator which has a U-Net \cite{U-NET10.1007/978-3-319-24574-4_28} architecture and the Discriminator is a PatchGAN \cite{Pix2Pix8100115} classifier. The pix2pix model not only learns the mapping from input to output image, but also constructs a loss function to train this mapping. Interestingly, unlike regular GANs, there is no random noise vector input to the pix2pix Generator. Instead, the Generator learns a mapping from the input image $x$ to the output image $G(x)$. The objective or the loss function for the Discriminator is the traditional adversarial loss function. The Generator on the other hand is trained using the adversarial loss along with the $L_1$ or pixel distance loss between the generated image and the real or target image. The $L_1$ loss encourages the generated image for a particular input to remain as similar as possible to the corresponding output real or ground truth image. This leads to faster convergence and more stable training. The loss function for conditional GAN is given by

\begin{equation}
\begin{aligned}
\mathcal{L}_{c G A N}(G, D)=& \mathbb{E}_{x, y}[\log D(x, y)]+
& \mathbb{E}_{x, z}[\log (1-D(x, G(x, z))]
\end{aligned}
\end{equation}

The $L_1$ or pixel distance loss is given by
\begin{equation}
    \mathcal{L}_{L 1}(G)=\mathbb{E}_{x, y, z}\left[\|y-G(x, z)\|_{1}\right]
\end{equation}

The final loss function is given by
\begin{equation}
    \arg \min _{G} \max _{D} \mathcal{L}_{c G A N}(G, D)+\lambda \mathcal{L}_{L 1}(G)
\end{equation}
where $\lambda$ is the weighting hyper-parameter coefficient. Pix2PixHD\cite{wang2018pix2pixHD} is an improved version of the Pix2Pix algorithm. The primary goal of Pix2PixHD is to produce high-resolution images and perform semantic manipulation. To do this the authors introduced multi-scale Generators and Discriminators and combined the cGANs and feature matching loss function.
The training set consists of pairs of corresponding images ${(s_i,x_i}$, where $s_i$ is a semantic label map and $x_i$ is a corresponding natural image. The cGAN loss function is given by,
\begin{equation}
    \mathbb{E}_{(\mathbf{s}, \mathbf{x})}[\log D(\mathbf{s}, \mathbf{x})]+\mathbb{E}_{\mathbf{s}}[\log (1-D(\mathbf{s}, G(\mathbf{s}))]
\end{equation}
The ith-layer feature extractor of Discriminator $D_k$ as ${D_k}^{(i)}$(from input to the $i$th layer of $D_k$). The feature matching loss $\mathcal{L}_{FM}(G, D_k)$ is given by
\begin{equation}
    \mathcal{L}_{\mathrm{FM}}\left(G, D_{k}\right)=\mathbb{E}_{(\mathbf{s}, \mathbf{x})} \sum_{i=1}^{T} \frac{1}{N_{i}}\left[\left\|D_{k}^{(i)}(\mathbf{s}, \mathbf{x})-D_{k}^{(i)}(\mathbf{s}, G(\mathbf{s}))\right\|_{1}\right]
\end{equation}
where $T$ is the total number of layers and $N_i$ denotes the number of elements in each layer. The objective function of pix2pixHD is given by
\begin{equation}
    \min _{G}\left(\left(\max _{D_{1}, D_{2}, D_{3}} \sum_{k=1,2,3} \mathcal{L}_{\mathrm{GAN}}\left(G, D_{k}\right)\right)+\lambda \sum_{k=1,2,3} \mathcal{L}_{\mathrm{FM}}\left(G, D_{k}\right)\right)
\end{equation}

\begin{figure}[h!]
    \centering
    \includegraphics[width=7cm , height=3cm]{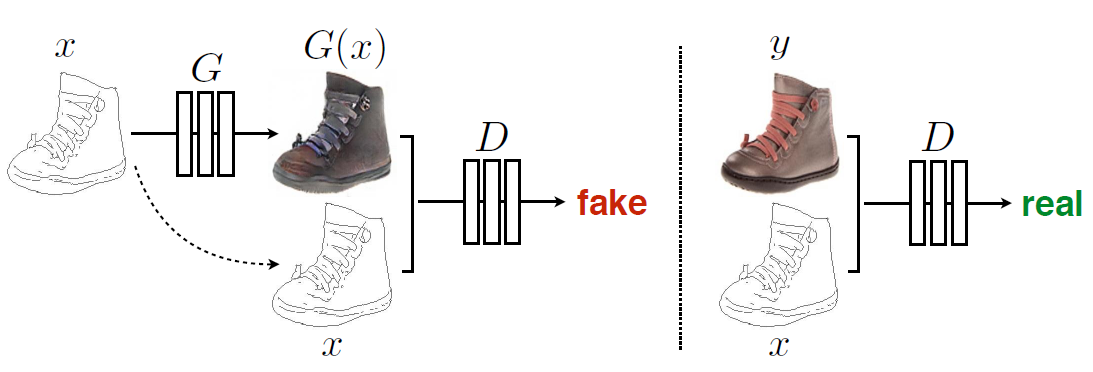}
    \caption{Using pix2pix to map edges to color images\cite{Pix2Pix8100115}. D, the Discriminator, learns to distinguish between fake (Generator-generated) and actual (edge, photo) tuples. G, the Generator, learns how to deceive the Discriminator. In contrast to an unconditional GAN, the Generator and Discriminator both look at the input edge map.}
    \label{fig:pix2pix}
\end{figure}

\subsection{Unpaired Image-to-Image Translation using Cycle-Consistent Adversarial Networks(CycleGAN)}
One fatal flaw of pix2pix is that it requires paired images for training and thus cannot be used for unpaired data which do not have input and output pairs. CycleGAN\cite{CycleGAN2017} addresses the issue by introducing a cycle consistency loss that tries to preserve the original image after a cycle of translation and reverse translation. Matching pairs of images are no longer required for training in this formulation. CycleGAN uses two Generators and two Discriminators. The Generator G is used to convert images from the X to the Y domain. The Generator F, on the other hand, converts images from Y to X. ($G : X \rightarrow Y$, $F : Y \rightarrow X$). The Discriminator $D_Y$ distinguishes $y$ from $G(x)$ and the Discriminator $D_X$ distinguishes $x$ from $F(y)$. The adversarial loss is applied to both the mapping functions. For the mapping function $G : X \rightarrow Y$ and its Discriminator $D_Y$ , the objective function is given by

\begin{equation}
\begin{aligned}
\mathcal{L}_{\mathrm{GAN}}\left(G, D_{Y}, X, Y\right) &=\mathbb{E}_{y \sim p_{\text {data }}(y)}\left[\log D_{Y}(y)\right] \\
&+\mathbb{E}_{x \sim p_{\text {data }}(x)}\left[\log \left(1-D_{Y}(G(x))\right]\right.
\end{aligned}
\end{equation}
The authors argued that the adversarial losses alone cannot guarantee that the learned function can map an individual input $x_i$ to a desired output $y_i$ as it leaves the model under-constrained. The authors therefore used the cycle consistency loss such that the learned mapping is cycle-consistent. It is based on the assumption that if you convert an image from one domain to the other and back again by feeding it through both Generators in sequence, you should get something similar to what you put in. 
Forward cycle consistency is represented as $x \rightarrow G(x) \rightarrow F(G(x)) \approx x$
and the backward cycle consistency as $y \rightarrow F(y) \rightarrow G(F(y)) \approx y$. The cycle consistency loss is given by
\begin{equation}
\begin{aligned}
\mathcal{L}_{\text {cyc }}(G, F) &=\mathbb{E}_{x \sim p_{\text {data }}(x)}\left[\|F(G(x))-x\|_{1}\right] \\
&+\mathbb{E}_{y \sim p_{\text {data }}(y)}\left[\|G(F(y))-y\|_{1}\right]
\end{aligned}
\end{equation}

The final full objective is given by
\begin{equation}
\begin{aligned}
\mathcal{L}\left(G, F, D_{X}, D_{Y}\right) &=\mathcal{L}_{\mathrm{GAN}}\left(G, D_{Y}, X, Y\right) \\
&+\mathcal{L}_{\mathrm{GAN}}\left(F, D_{X}, Y, X\right) \\
&+\lambda \mathcal{L}_{\mathrm{cyc}}(G, F),
\end{aligned}
\end{equation}
where $\lambda$  controls the relative importance of the two objectives. 
\begin{equation}
G^{*}, F^{*}=\arg \min _{G, F} \max _{D_{x}, D_{Y}} \mathcal{L}\left(G, F, D_{X}, D_{Y}\right)
\end{equation}

\begin{figure}[h!]
    \centering
    \includegraphics[width=\textwidth]{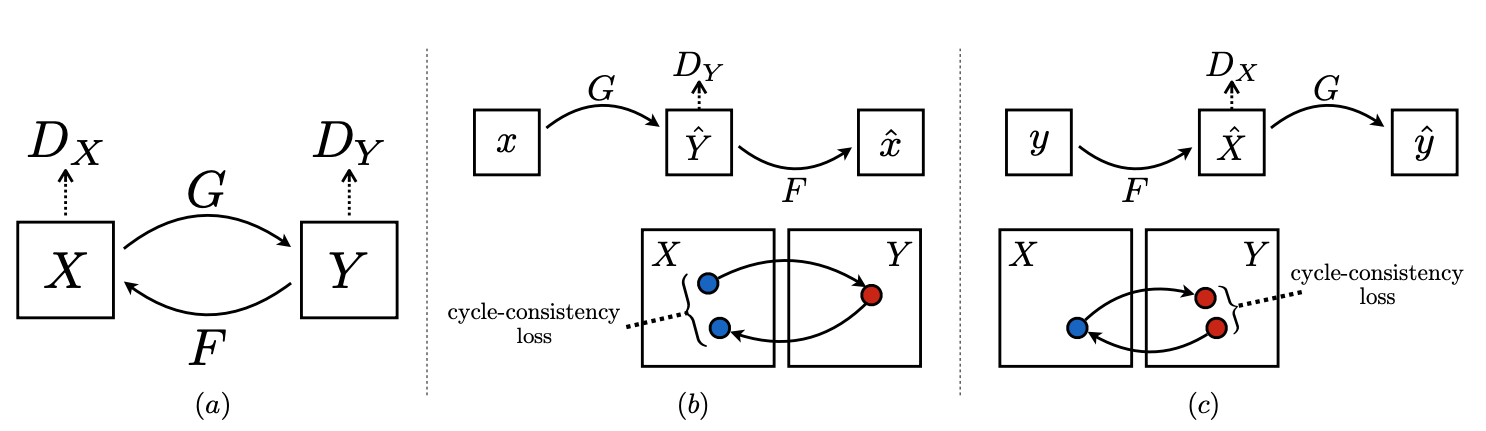}
    \caption{CycleGAN \cite{CycleGAN2017} (a)  two Generator mapping functions $G : X \rightarrow Y$ and $F : Y \rightarrow X$, and two Discriminators $D_Y$ and $D_X$. (b)  forward cycle-consistency loss: $x \rightarrow G(x) \rightarrow F(G(x)) \approx x$ and (c) backward cycle consistency loss: $y \rightarrow F(y) \rightarrow G(F(y)) \approx y$.}
    \label{fig:CycleGAN}
\end{figure}

\subsection{A Style-Based Generator Architecture for Generative Adversarial Networks(StyleGAN)}
\begin{figure}[h!]
    \centering
    \includegraphics[width=5cm, height=7cm]{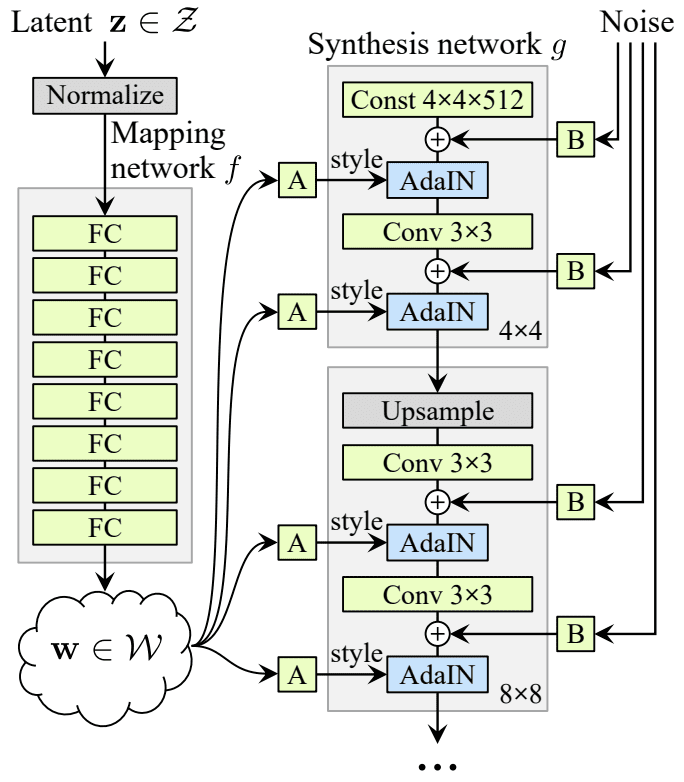}
    \caption{StyleGAN Generator \cite{Karras_2020_CVPR}}
    \label{fig:StyleGAN}
\end{figure}

The primary goal of StyleGAN\cite{Karras_2020_CVPR} is to produce high quality, high resolution facial images that are diverse in nature and provide control over the style of generated synthetic images. StyleGAN is an extension of the ProGAN\cite{PROGAN:DBLP:journals/corr/abs-1710-10196} model which uses the progressive growing approach for synthesizing high resolution and high quality images via the incremental (gradual) growing of the Discriminator and the Generator networks during the training process. It's important to note that StyleGAN changes affect only the Generator network, which means they only affect the generative process. The Discriminator and loss function, which are both the same as in a traditional GAN, have not been altered. The upgraded Generator includes several additions to the ProGAN’s Generators which is shown in Figure \ref{fig:StyleGAN}. and are described below:

\begin{itemize}
    \item \textbf{Baseline Progressive GAN:} The authors use the Progressive GAN(ProGAN\cite{PROGAN:DBLP:journals/corr/abs-1710-10196}) as their baseline from which they inherit the network architecture and some of the hyperparameters. 
    
    \item \textbf{Bi-linear up/down sampling:} The ProGAN model used the nearest neighbor up/down sampling but the authors of StyleGAN used bi-linear sampling layers for both the Generator and the Discriminator.
    
    \item \textbf{Mapping Network, Style Network and AdaIN:} Instead of feeding in the noise vector $z$ directly into the Generator, it goes through a mapping network to get an intermediate noise vector $w$, say. 
    The output of the mapping network ($w$) is then passed through a learned affine transformation (A) before passing into the synthesis network through the  Adaptive Instance Normalization\cite{8237429} or AdaIN module. In Figure ``A'' stands for a learned affine transform.  The AdaIN module transfers the encoded information, created by the Mapping Network after the affine transformation, which is incorporated into each block of the Generator model after the convolutional layers. The AdaIN module begins by converting the output of the feature map to a standard Gaussian and then adding the style vector as a bias term.  The mapping network $f$ is a standard deep neural network which is comprised of 8 fully connected layers and the synthesis network $g$ consists of 18 layers.
    
    \item \textbf{Remove traditional input:} 
    Most models, including ProGAN, utilize random input to generate the Generator's initial image. However, the StyleGAN authors found that the image features are controlled by $w$ and the AdaIN. As a result, they simplify the architecture by eliminating the traditional input layer and begin image synthesis with a learned constant tensor.
    
    \item \textbf{Add noise inputs:} Before evaluating the nonlinearity, Gaussian noise is added after each convolution. In Figure 7. ``B'' is the learned scaling factor applied per channel to the noise input.
    
    \item \textbf{Mixing regularization:} The authors also introduced a novel regularization method to reduce neighbouring styles correlation and have more fine grained control over the generated images. Instead of passing just one latent vector, $z$, through the mapping network as input and getting one vector, $w$, as output, mixing regularisation passes two latent vectors, $z_1$ and $z_2$, through the mapping vector and gets two vectors, $w_1$ and $w_2$. The use of $w_1$ and $w_2$ is completely randomized for every iteration this technique prevents the network from assuming that styles adjacent to each other correlate. 
\end{itemize}

\subsection{Recurrent GAN (RGAN) and Recurrent Conditional GAN (RCGAN)}
Besides generating synthetic images, GAN can also generate sequential data\cite{mogren2016c, esteban2017real}. Instead of modeling the data distribution in the original feature space, the generative model for time-series data also captures the conditional distribution $P(X_t|X_{1:t-1})$ given historical data. The main difference in architecture between Recurrent GAN and the traditional GAN is that we replace the DNNs/ CNNs with Recurrent Neural Networks (RNNs) in both Generator and Discriminator. Here, the RNNs can be any variants of RNN, such as Long short-term memory (LSTM) and  Gated Recurrent Unit (GRU), which captures the temporal dependency in input data. In the case of Recurrent Conditional GAN (RCGAN), both Generator and Discriminator are conditioned on some auxiliary information. Many experiments\cite{esteban2017real} show that RGAN and RCGAN are able to effectively generate realistic time-series synthetic data. 

\begin{figure}[h!]
    \centering
    \includegraphics[width=10cm,height=4cm]{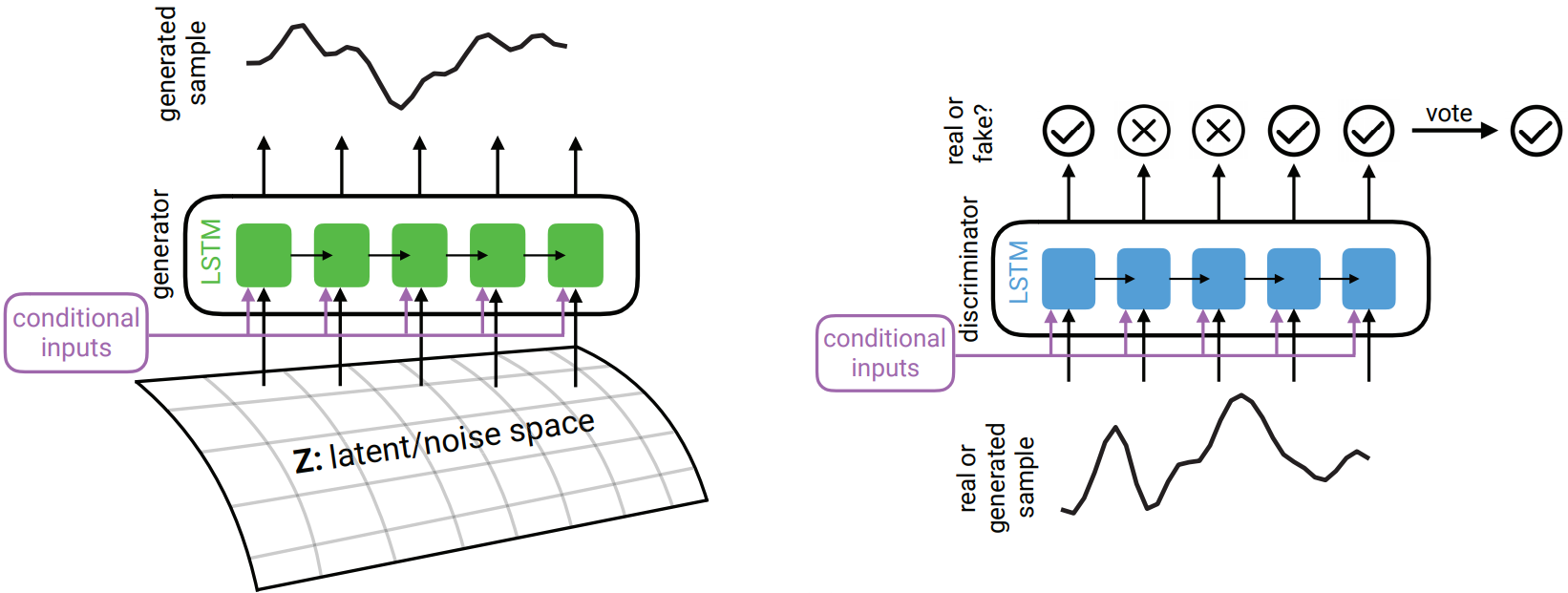}
    \caption{RGAN and RCGAN \cite{esteban2017real}} 
    \label{fig:RGAN-and-RCGAN}
\end{figure}

In Figure \ref{fig:RGAN-and-RCGAN}, we illustrate the architecture of RGAN and RCGAN. The Generator RNN takes the random noise at each time step to generate the synthetic sequence. Then, the Discriminator RNN works as a classifier to distinguish whether the input is real or fake. Condition inputs are concatenated to the sequential inputs of both the Generator and Discriminator if it is an RCGAN. Similar to GAN, the Discriminator in RGAN minimize the cross-entropy loss between the generated data and the real data. The Discriminator loss is formulated as follows.

\begin{equation}
D_{loss}(X_n, y_n) = -CE({RNN}_{D}(X_n), y_n)
\end{equation}

Where $X_n$ ($X_n \in \mathbb{R}^{T \times d}$) and $y_n$ ($y_n \in \{1, 0\}^T$) are the input and output of the Discriminator with sequence length $T$ and feature size $d$. $y_n$ is a vector of 1s for real sequence and 0s for synthetic sequence. $CE(\cdot)$ is the average cross-entropy function and ${RNN}_{D}(\cdot)$ is the RNN in Discriminator. The Generator loss is formulated below.

\begin{equation}
G_{loss}(Z_n) = D_{loss}({RNN}_{G}(Z_n), 1)= -CE({RNN}_D({RNN}_G(Z_n)), 1)
\end{equation}

Here, $Z_n$ is a random noise vector with $Z_n \in \mathbb{R}^{T \times m}$. In the case of RCGAN, the inputs of both Generator and Discriminator also concatenate the conditional information $c_n$ at each time step.

\subsection{Time-series GAN (TimeGAN)}
Recently, a novel GAN framework preserving temporal dynamics called Time-Series GAN (TimeGAN) \cite{yoon2019time} is proposed. Besides minimizing the unsupervised adversarial loss in the traditional GAN learning procedure, (1) TimeGAN introduces a stepwise supervised loss using the original inputs as supervision, which explicitly encourages the model to capture the stepwise conditional distributions in the data. (2) TimeGAN introduces an autoencoder network to learn the mapping from feature space and embedding/latent space, which reduces the dimensionality of the adversarial learning space. (3) To minimize the supervised loss, jointly training on both the autoencoder and Generator is deployed, which forces the model to be conditioned on the embedding to learn temporal relationship. TimeGAN framework not only captures the distributions of features in each time step but also capture the complex temporal dynamics of features across time.

\begin{figure}[h!]
    \centering
    \includegraphics[width=10cm,height=5cm]{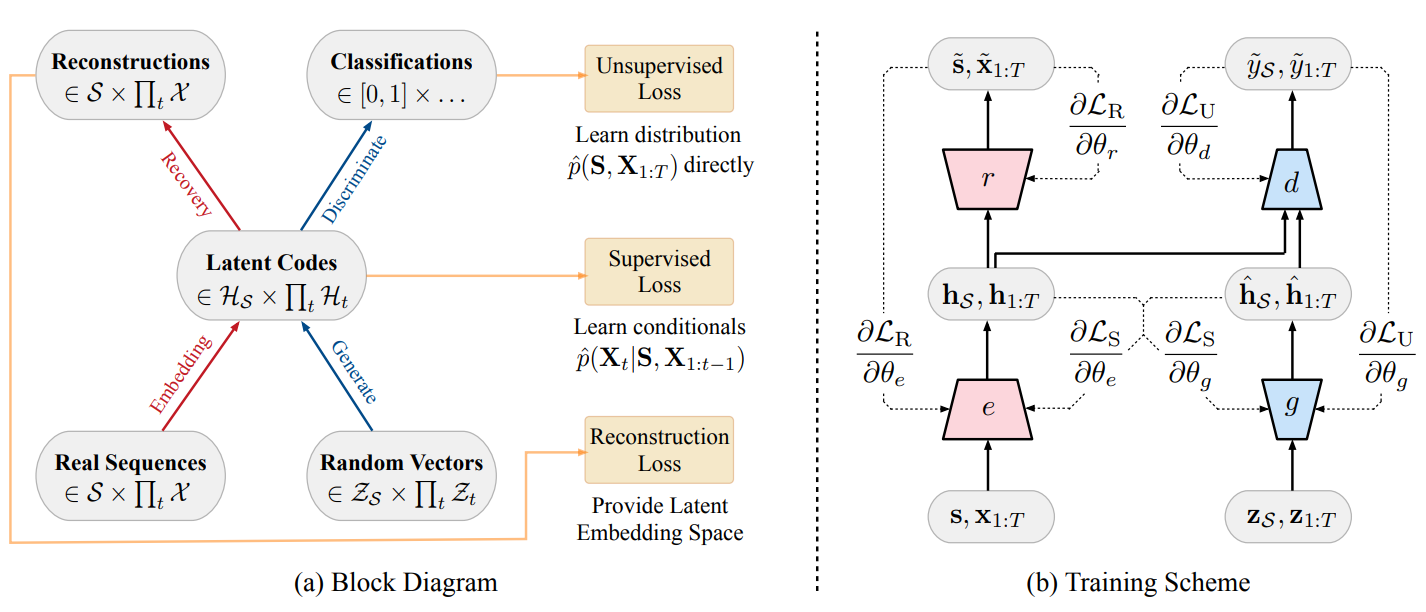}
    \caption{(a) Block Diagram of Four Key Components in TimeGAN. (b) Training scheme: solid lines and dashed lines represent forward propagation paths and backpropagation paths respectively.\cite{yoon2019time}}
    \label{fig:TimeGAN}
\end{figure}

The TimeGAN consists of four important parts, embedding function, recovery function, Generator and Discriminator in Figure \ref{fig:TimeGAN}. First, the autoencoder (first two parts) learns the latent representation from the inputs sequence. Then, the adversarial model (latter two parts) trains jointly on the latent space to generate the synthetic sequence with temporal dynamics by minimizing both the unsupervised loss and supervised loss.

\begin{table}[h!]
    \caption{Application of common GAN models}
    \begin{tabular}{p{0.4\textwidth}p{0.55\textwidth}}
        \hline
        \textbf{Common GAN variants and extensions} &
        \textbf{Application area}  \\
        \hline
        DCGAN\cite{radford2016unsupervised} & Image Generation \\ 
        cGAN\cite{cGANmirza2014conditional} & Semi supervised conditional Image Generation \\ 
        InfoGAN\cite{DBLP:journals/corr/ChenDHSSA16} &  Unsupervised learning of interpretable and disentangled representations\\ 
        StackGAN\cite{StackGAN8237891} & Image generation based on text inputs\\ 
         Pix2Pix\cite{Pix2Pix8100115} &  Image to image translation\\ 
         CycleGAN\cite{CycleGAN2017} &  Unpaired Image to image translation\\ 
         StyleGAN\cite{Karras_2020_CVPR} &  High resolution facial image generation that are diverse in nature\\ 
         RGAN, RCGAN\cite{esteban2017real} & Synthetic medical time series data generation \\
         TimeGAN\cite{yoon2019time} & Realistic time-series data generation \\
        \hline
    \end{tabular}
    
    \label{tab:GANs}
\end{table}

\section{GAN evaluation metrics}
One of the most difficult aspects of GAN training is assessing their performance, or determining how well a model approximates a data distribution. In terms of theory and applications, significant progress has been made, with a large number of GAN variants now available. However there has been relatively little effort put into evaluating GANs, and there are still gaps in quantitative assessment methods.
In this section, we present the relevant and popular metrics which are used to evaluate the performance of GANs. 

\begin{enumerate}
    \item \textbf{Inception Score (IS):} IS was proposed by Salimans et al.\cite{IS_NIPS2016_8a3363ab} and it employs the pre-trained InceptionNet\cite{Inception_net_7780677} trained on ImageNet\cite{ImageNet_5206848} to capture the desired properties of generated samples. The average Kullback–Leibler or KL divergence\cite{KL_Joyce2011} between the conditional label distribution $p(y \mid \mathbf{x})$ of samples and the marginal distribution $p(y)$ calculated from all samples is measured by IS. The goal of IS is to assess two characteristics for a set of generated images: image quality (which evaluates whether images have meaningful objects in them) and image diversity. Thus IS favors a low entropy of $p(y \mid \mathbf{x})$ but a high entropy of $p(y)$. IS can be expressed as:
    \begin{equation}
    \exp \left(\mathbb{E}_{\mathbf{x}}[\mathbb{K} \mathbb{L}(p(\mathrm{y} \mid \mathbf{x}) \| p(\mathrm{y}))]\right)
    \end{equation}
    
    A higher IS indicates that the generative model is capable of producing high-quality samples that are also diverse.
    
    \item \textbf{Modified Inception Score (m-IS):} In its original form, Inception Score assigns models that produce a low entropy class conditional distribution $p(y \mid \mathbf{x})$ with a higher score overall generated data. It is, however, desirable to have diversity within a category of samples. To characterize this diversity, Gurumurthy et al.\cite{mod_IS_8100008} proposed a modified version of inception-score which incorporates a cross-entropy style score $-p\left(y \mid \mathbf{x}_{i}\right) \log \left(p\left(y \mid \mathbf{x}_{j}\right)\right)$ where $x_{j}s$ are samples of the same class as $x_i$ as per the outputs of the trained inception model. The modified IS can be defined as,
    \begin{equation}
    \exp \left(\mathbb{E}_{\mathbf{x}_{i}}\left[\mathbb{E}_{\mathbf{x}_{j}}\left[\left(\mathbb{K} \mathbb{L}\left(P\left(y \mid \mathbf{x}_{i}\right) \| P\left(y \mid \mathbf{x}_{j}\right)\right)\right]\right]\right)\right.
    \end{equation}
    The m-IS is calculated per-class and then averaged across all classes. m-IS can be thought of as a proxy for assessing both intra-class sample diversity and sample quality.
    
    \item \textbf{Mode Score(MS):} The MODE score introduced by Che et al.\cite{ModeScoreChe2017ModeRG} is an improved version of the IS that addresses one of the IS's major flaws: it ignores the prior distribution of ground truth labels. In contrast to IS, MS can measure the difference between the real and generated distributions.
    \begin{equation}
    \exp \left(\mathbb{E}_{\mathbf{x}}\left[\mathbb{K} \mathbb{L}\left(p(y \mid \mathbf{x}) \| p\left(y^{\text {train }}\right)\right)\right]-\mathbb{K} \mathbb{L}\left(p(y) \| p\left(y^{\text {train }}\right)\right)\right)
    \end{equation}
    where $p\left(\mathrm{y}^{\text {train }}\right)$ is the empirical distribution of labels computed from training data. According to the author's evaluation, the MODE score successfully measures two important aspects of generative models, namely variety and visual quality.
    
    \item \textbf{Frechet Inception Distance (FID):} Proposed by Heusel et al.\cite{FID10.5555/3295222.3295408},
    the Frechet Inception Distance score determines how far feature vectors calculated for real and generated images differ. A specific layer of the InceptionNet\cite{Inception_net_7780677} model is used by the FID score to capture and embed features of an input image. The embeddings are summarized as a multivariate Gaussian by calculating the mean and covariance for both the generated data and the real data. The Fréchet distance (or Wasserstein-2 distance) between these two Gaussians is then used to quantify the quality of the generated samples.
    \begin{equation}
    F I D(r, g)=\left\|\mu_{r}-\mu_{g}\right\|_{2}^{2}+\operatorname{Tr}\left(\Sigma_{r}+\Sigma_{g}-2\left(\Sigma_{r} \Sigma_{g}\right)^{\frac{1}{2}}\right)
    \end{equation}
    where $\left(\mu_{g}, \Sigma_{g}\right)$ and  $\left(\mu_{r}, \Sigma_{r}\right)$ represent the empirical mean and empirical covariance of the generated and real data disctibutions respectively. Smaller distances between synthetic(model generated) and real data distributions are indicated by a lower FID.

    \item \textbf{Image Quality Measures:}
    Below we describe some commonly used image quality assessment measures used to compare GAN generated data with the real data.
        \begin{enumerate}
            \item Structural similarity index measure (SSIM)\cite{SSIM1284395} is a method for quantifying the similarity between two images. SSIM tries to model the perceived change in the image's structural information. The SSIM value varies between -1 and 1, where a value of 1 shows perfect similarity. Multi-Scale SSIM or MS-SSIM\cite{MSSSIM1292216} is a multiscale version of SSIM that allows for more flexibility in incorporating image resolution and viewing conditions than a single scale approach. MS-SSIM ranges between 0 (low similarity) and 1 (high similarity).
            \item Peak Signal-to-Noise Ratio or PSNR compares the quality of a generated image to its corresponding real image by measuring the peak signal-to-noise ratio of two monochromatic images. For example evaluation of conditional GANs or cGANs\cite{cGANmirza2014conditional} Higher PSNR (in db) indicates better quality of the generated image.
            \item Sharpness Difference (SD) represents the difference in clarity between the generated and the real image. The larger the value is, the smaller the difference in sharpness between the images is and the closer the generated image is to the real image.
            \end{enumerate}

    \item \textbf{Evaluation Metrics for Time-Series/Sequence Inputs:}
    To evaluate quality of generating synthetic sequence data is very challenging. For example, the  Intensive Care Unit (ICU) signal looks completely random to a non-medical export\cite{esteban2017real}. The researchers evaluate the quality of synthetic sequential data mainly focusing on the following three different aspects. (1) Diversity -- the synthetic data should be generated from the same distribution of real data. (2) Fidelity -- the synthetic data should be indistinguishable from the real data. (3) Usability --- the synthetic data should be good enough to be used as the train/test dataset.\cite{yoon2019time}
        \begin{enumerate}
        
            \item t-SNE and PCA \cite{yoon2019time} are both commonly used visualization tools for analyzing both the original and synthetic sequence datasets. They flatten the dataset across the temporal dimension so that the dataset can be plotted in the 2D plane. They measure how closely the distribution of generated samples resembles that of the original in 2-dimensional space.
            
            \item Discriminative Score \cite{yoon2019time} evaluates how difficult for a binary classifier to distinguish between the real (original) dataset and the fake (generated) dataset. It is challenging for the classifier to classify if the synthetic data and the original are drawn from the same distribution.
            
            \item Maximum Mean Discrepancy (MMD) \cite{gretton2008kernel, esteban2017real} learns the distribution of the real data. The maximum mean discrepancy method has been proposed to distinguish whether the synthetic data and the real data are from the same distribution by computing the squared difference of the statistics between real and synthetic samples ($MMD^2$). The unbiased $MMD^2$ can be denoted as following where the inner production between functions is replaced with a kernel function $K$.
            
            \begin{equation}
                \widehat{MMD^2} = \frac{1}{n(n-1)}\sum_{i=1}^{n}\sum_{j \neq i}^{n}K(x_i, x_j) - \frac{2}{mn}\sum_{i=1}^{n}\sum_{j=1}^{m}K(x_i, y_j) + \frac{1}{m(m-1)}\sum_{i=1}^{m}\sum_{j \neq i}^{m}K(y_i, y_j)
            \end{equation}
            
            A suitable kernel function for the time-series data is vital. The authors treat the time series as vectors for comparison and select the radial basis
            function (RBF) as the kernel function which is $K(x, y) = exp(- \left\Vert{x-y}\right\Vert^2/(2\sigma^2))$. To select an appropriate kernel bandwidth $\sigma$, the estimator of the t-statistic of the power of the MMD test between two distributions $\widehat{t} = \frac{\widehat{MMD}^2}{\sqrt{\widehat{V}}}$ is maximised. The authors split a validation set during training to tune the parameter. The result shows that $MMD^2$ is more informative than either Generator or Discriminator loss, and correlates well with quality as assessed by visualising\cite{esteban2017real}.

            \item Earth Mover Distance (EMD) \cite{wiese2020quant, villani2009optimal} is a measure of the distance between two probability distributions over a region. It describes how much probability mass has to be moved to transform $P^h$ into $P^g$ where $P^h$ denotes the historical distribution and $P^g$ is the generated distribution. The EMD is defined by
            
            \begin{equation}
            EMD(P^h, P^g) = \underset{\pi\in \prod_{}^{}(P^h, P^g)}{inf} E_{(X,Y)~\pi}[\left\Vert{X-Y}\right\Vert]
            \end{equation}
            
            where $\prod_{}^{}(P^h, P^g)$ denotes the set of all joint probability distributions with marginals $P^h$ and $P^g$.

            
            
            
            

            \item AutoCorrelation Function (ACF) Score \cite{wiese2020quant} describes the coefficient of correlation between historical and the generated time series. Let $r_{1:T}$ denote the historical log percentage change series and $\{r_{1:\widetilde{T},\theta}^{(1)},...,r_{1:\widetilde{T},\theta}^{(M)}\}$ a set of generated log percentage change paths of length $\widetilde{T} \in N$. The autocorrelation is calculated with the time lag $tau$ and the series $r_{1:T}$ and measures the correlation of the lagged time series with the series itself
            
            \begin{equation}
            C(\tau;r) = Corr(r_{t+\tau}, r_t)
            \end{equation}
            
            The ACF($f$) score is computed for a function $f: R \rightarrow R$ as
            
            \begin{equation}
            ACF(f) := \parallel C(f(r_{1:T})) - \frac{1}{M} \sum_{i=1}^{M} C(f(r_{1:T,\theta}^{(i)})) \parallel _2
            \end{equation}
            
            where $C: R^T \rightarrow [-1,1]^S:r_{1:T} \mapsto (C(1;r),...,C(S;r))$.

            \item Leverage Effect Score \cite{wiese2020quant} provides a comparison of the historical and the generated time dependence. The leverage effect for lag $\tau$ is measured using the correlation of the lagged, squared log percentage changes and the log percentage changes themselves.
            
            \begin{equation}
            \ell (\tau;r) = Corr(r_{t+\tau}^2, r_t)
            \end{equation}
            
            The leverage effect score is defined by
            
            \begin{equation}
            \parallel L(r_{1:T} = \frac{1}{M} \sum_{i=1}^{M} L(r_{1:\widetilde{T},\theta}^{(i)})) \parallel_2
            \end{equation}
            
             where $L: R^T \rightarrow [-1,1]^S: r_{1;r} \mapsto (\ell(1;r),...,\ell(S;r))$.
        \end{enumerate}

\end{enumerate}

\begin{table}[h!]
\caption{Summary of relevant and popular GAN evaluation metrics}
\begin{tabular}{p{0.45\textwidth}p{0.5\textwidth}}
\hline
\textbf{Quantitative metrics} & \textbf{Description}       \\ 
\hline
Inception Score (IS) \cite{IS_NIPS2016_8a3363ab} & Measures the KL-Divergence between the conditional and marginal label distributions over the data.  \\ 
Modified Inception Score (m-IS) \cite{mod_IS_8100008} & Incorporates a cross-entropy style score to promote diversity within a category of samples.\\
Mode Score (MS) \cite{ModeScoreChe2017ModeRG} & Improved veersion of IS and takes into account the  prior distribution of ground truth labels. \\
Fréchet Inception Distance (FID) \cite{FID10.5555/3295222.3295408} & Evaluates the Fréchet distance or the Wasserstein-2 distance between the multi-variate Gaussians fitted to data embedded into a feature space.\\
Structural Similarity Index Measure (SSIM\cite{SSIM1284395}),
Peak signal-to-noise ratio (PSNR),
Multiscale SSIM  (MS-SSIM\cite{MSSSIM1292216}) and
Sharpness Difference (SD) & Evaluate and assess the quality of generated images. \\
Maximum Mean Discrepancy (MMD\cite{gretton2008kernel, mogren2016c}), Earth  Mover Distance (EMD \cite{wiese2020quant, villani2009optimal}), DY Metric\cite{wiese2020quant}, ACF  Score\cite{wiese2020quant}, Leverage  Effect  Score\cite{wiese2020quant} & Evaluate the quality of generated sequence data. \\
\hline
\end{tabular}
\label{tab:metrics}
\end{table}

\section{GAN applications}
GANs are by far the most widely used generative models and they are immensely powerful for the generation of realistic synthetic data samples. In this section, we will go over the wide array of domains in which Generative Adversarial Networks (GANs) are being applied. Specifically, we will present the use of GANs in Image processing, Video generation and prediction, Medical and Healthcare, Biology, Astronomy, Remote Sensing, Material Science, Finance, Fashion Design, Sports and Music.

\subsection{Image processing}
GANs are quite prolific when it comes to specific image processing related tasks like image super-resolution, image editing, high resolution face generation, facial attribute manipulation to name a few.

\begin{itemize}
    \item \textbf{Image super-resolution:} Image super-resolution refers to the process of transforming low resolution images to high resolution images. SRGAN\cite{ledig2017photorealistic} is the first image super-resolution framework capable of inferring photo-realistic natural images for 4x up-scaling factors. Several other super resolution frameworks(\cite{TSRGANapp10051729}, \cite{ESRGAN10.1007/978-3-030-11021-5_5}) have also been developed to produce better results. Best-Buddy GANs\cite{li2021bestbuddy} developed recently is used for single image super-resolution (SISR) task along with previous works(\cite{Zhu2020}, \cite{Perception10.1007/978-3-030-11021-5_7}).
    \item \textbf{Image editing:} Image editing involves removing or modifying some aspects of an image. For example, images captured during bad weather or heavy rain lack visual quality and thus will require manual intervention to either remove anomalies such as raindrops or dust particles that reduce image quality. The authors of ID-CGAN\cite{De-Raining8727938}  used GANs to address the problem of single image de-raining. Image modification could involve modifying or changing some aspects of an image such as changing the hair color, adding a smile, etc. which was demonstrated by (\cite{perarnau2016invertible}, \cite{zhuang2021enjoy}). 
    
    \item \textbf{High resolution face image generation:} High resolution facial image generation is one other area of image processing where GANs have excelled. Face generation and attribute manipulation using GANs can be broadly classified into the creation of entire synthetic faces, face features or attribute manipulation and face component transformation.
    \begin{itemize}
        \item \textbf{Synthetic face generation:} Synthetic face generation refers to the creation of synthetic images of the face of people who do not exist in real life. ProGAN\cite{PROGAN:DBLP:journals/corr/abs-1710-10196}, described in the previous section demonstrated the generation of realistic looking images of human faces. Since then there have been several works which use GANs for facial image generation(\cite{Generation9067742}, \cite{generationZHANG2020102719}). StyleGAN\cite{StyleGANDBLP:journals/corr/abs-1812-04948} which is a unique generative adversarial network introduced by Nvidia researchers in December 2018. The primary goal of StyleGAN is to generate high quality face images that are also diverse in nature. To achieve this, the authors used techniques such as using a noise mapping network, adaptive instance normalization and progressive growing similar to ProGAN to produce very high resolution images.
        \item \textbf{Face features or attribute manipulation:} Face attribute manipulation includes facial pose and expression transformation. The authors of PosIX-GAN\cite{10.1007/978-3-030-11015-4_31} trained their model to generate high quality face images with 9 different pose variations when given a face image in any arbitrary pose as input. DECGAN\cite{DECGAN8483579} authors used Double Encoder Conditional GAN to perform facial expression synthesis. Expression Conditional GAN (ECGAN)\cite{ECGANDBLP:journals/corr/abs-1905-05416} can learn the mapping from one image domain to another and the authors were able to control specific facial expressions by the conditional attribute label.
    \item \textbf{Face component transformation:} Face component transformation deals with altering the face style(hair color and style) or adding accessories such as eye glasses. The authors of DiscoGAN\cite{DBLP:journals/corr/KimCKLK17} were able to change hair color and the authors of StarGAN\cite{StarGANDBLP:journals/corr/abs-1711-09020} were able to perform multi-domain image translations. BeautyGAN\cite{BeautyGAN10.1145/3240508.3240618} can be used to translate the makeup style from a given reference makeup face image to another non-makeup one while preserving face identity. The authors of InfoGAN\cite{DBLP:journals/corr/ChenDHSSA16} trained their model to learn disentangled representations in an unsupervised manner and can modify facial components such as adding or removing eyeglasses and changing hairstyles. GANs can also be applied to image inpainting where the task is of reconstructing missing regions in an image. In this regard GANs have been used(\cite{8100211}, \cite{8578675}) to perform the task.
    \end{itemize}
\end{itemize}

\begin{table}[h!]
    \caption{Applications in Image processing}
    \begin{tabular}{p{0.5\textwidth}p{0.45\textwidth}}
        \hline
        \textbf{Image processing application} &
        \textbf{GAN models}  \\
        \hline
        Image super-resolution & SRGAN \cite{ledig2017photorealistic}, TSRGAN \cite{TSRGANapp10051729}, ESRGAN \cite{ESRGAN10.1007/978-3-030-11021-5_5}, Best-BuddyGAN \cite{li2021bestbuddy}, GMGAN \cite{Zhu2020}, PESR \cite{Perception10.1007/978-3-030-11021-5_7} \\ 
        Image editing and modifying & ID-CGAN \cite{De-Raining8727938}, IcGAN \cite{perarnau2016invertible} \\ 
        Synthetic face generation & ProGAN \cite{PROGAN:DBLP:journals/corr/abs-1710-10196}, StyleGAN \cite{Karras_2020_CVPR} \\ 
        Face features or attribute manipulation & PosIX-GAN \cite{10.1007/978-3-030-11015-4_31}, DECGAN \cite{DECGAN8483579}, ECGAN \cite{ECGANDBLP:journals/corr/abs-1905-05416} \\ 
        Face component transformation & DiscoGAN \cite{DBLP:journals/corr/KimCKLK17}, StarGAN \cite{StarGANDBLP:journals/corr/abs-1711-09020}, BeautyGAN \cite{BeautyGAN10.1145/3240508.3240618}, InfoGAN \cite{DBLP:journals/corr/ChenDHSSA16} \\
        \hline
    \end{tabular}
    \label{tab:table2}
\end{table}

\subsection{Video generation and prediction}
Synthesizing videos using GANs can be divided into three main categories. (a) Unconditional video generation (b) Conditional video generation (c) Video prediction

\subsubsection{Unconditional video generation}
For \textbf{unconditional video generation} the output of the GAN models is not conditioned on any input signals. Due to the lack of any information provided as a condition with the videos during the training phase, the output videos produced by these frameworks typically are low-quality in nature. 

The authors of VGAN\cite{VGAN10.5555/3157096.3157165} were the first to apply GANs for video generation. Their Generator consists of two CNN networks, one 3D spatio-temporal convolutional network to capture moving objects in the foreground, and the other is a 2D spatial convolutional model that captures the static background. The two independent outputs from the Generator are combined to create the generated video and fed to the Discriminator, to decide if it is real or fake. Temporal Generative Adversarial Nets (TGAN)\cite{TGAN} can learn representation from an unlabeled video dataset and generate a new video. TGAN  Generator consists of two sub Generators one of which is the temporal Generator and the other is the image Generator. The temporal Generator takes a single latent variable as input and produces a set of latent variables, each of which corresponds to a video frame. The image Generator creates a video from a set of latent variables. The Discriminator consists of three-dimensional convolutional layers. TGAN uses WGAN to provide stable training and meets the K-Lipschitz constraint. FTGAN\cite{FTGAN:DBLP:journals/corr/abs-1711-09618} consists of two GANs: FlowGAN and TextureGAN. FlowGAN network deals with motion, i.e. adds optical flow for representing the object motion
more effectively. The TextureGAN model is used to generate the texture that is conditioned on the previous FlowGAN result, to produce the required frames. Motion and Content decomposed Generative Adversarial Network or MoCoGAN\cite{Tulyakov:2018:MoCoGAN} uses a motion and content decomposed representation for video generation. MoCoGAN is made up of four sub-networks: a recurrent neural network, an image Generator, an image Discriminator, a video Discriminator, and a video Discriminator. Built on the BigGAN\cite{brock2019large} architecture, the Dual Video Discriminator GAN (DVD-GAN)\cite{clark2019adversarial} is a generative video model for high quality frame generation. DVD-GAN employs Recurrent Neural Network (RNN) units as well as a dual Discriminator architecture to deal with the spatial and temporal dimension.
\subsubsection{Conditional video generation}
In \textbf{conditional video generation} the output of the GAN models is conditioned on input signals such as text, audio or speech. We do not consider other conditioning techniques such as images to video, semantic map to videos and video to video as these can be considered to fall under the video prediction category which is described in section 4.5.2 below. 

For \textbf{text to video synthesis} the goal is to generate videos based on some conditional text. Li et al.\cite{li2017video} used  Variational Autoencoder (VAE)\cite{VAE_kingma2014autoencoding} and Generative Adversarial Network (GAN) for generating videos from text. Their model is made a conditional gist Generator (conditional VAE), a video Generator, and a video Discriminator. The initial image/gist is created by the conditional gist Generator, which is conditioned on encoded text. This gist serves as a general representation of the image, background color and object layout of the desired video. The video's content and motion are then generated using cGAN by conditioning both the gist and the text input. Temporal GANs conditioning on captions (TGANs-C)\cite{TGAN} uses a Bidirectional LSTM and LSTM based encoder to embed and obtain the representation of the input text. This output representation is then concatenated with a random noise vector and then given to the 
Generator, which is a 3D deconvolution network to generate synthesize realistic videos. The model has three Discriminators: The video Discriminator distinguishes real video from synthetic video and aligns video with the correct caption, the frame Discriminator determines whether each frame is real/fake and semantically matched/mismatched with the given caption, and the motion Discriminator exploits temporal coherence between consecutive frames. Balaji et al. proposed the Text-Filter conditioning Generative Adversarial Network (TFGAN)\cite{TFGANijcai2019-276}. TFGAN introduces a novel multi-scale text-conditioning technique in which text features are extracted from the encoded text and used to create convolution filters. Then, the convolution filters are input to the Discriminator network to learn good video-text associations in the GAN model. StoryGAN\cite{StoryGAN8953914} is based on a sequential conditional GAN framework whose task is to generates a sequence of images for each sentence in a given multi-sentence paragraph. The GAN model consists of $(i)$ Story Encoder, $(ii)$ A recurrent neural network (RNN) based Context Encoder, $(iii)$  An image Generator and $(iv)$ An image Discriminator and a story Discriminator.
BoGAN\cite{BoGAN9138828} maintains semantic matching between video and the corresponding language description at various levels and ensures coherence between consecutive frames. The authors used LSTM and 3D convolution based encoder decoder architecture to produce frames from embedding based on the input text. Region level semantic alignment module was proposed to encourage the Generator to take advantage of the semantic alignment between video and words on a local level. To maintain frame-level and video level coherence two Discriminators were used. Kim et al.\cite{TiVGAN9171240} came up with Text-to-Image-to-Video Generative Adversarial Network (TiVGAN) for text-based video generation. The key idea is to begin by learning text-to-single-image generation, then gradually increase the number of images produced, and repeat until the desired video length is reached.

\textbf{Speech to video synthesis} involves the task of generating synchronized video frames conditioned on an audio/speech input. Jalalifar et al.\cite{jalalifar2018speechdriven} used LSTM and CGAN for speech conditioned talking face generation. LSTM network learns to extract and predict facial landmark positions from audio features. Given the extracted set of landmarks, the cGAN then generates synchronized facial images with accurate lip sync. Vougioukas et al.\cite{Vougioukas_2019_CVPR_Workshops} used GANs for generating videos of a talking head. The generation of video frames is conditioned on a still image of a person and an audio clip containing speech and does not rely on extracting intermediate features. Their GAN architecture uses an RNN based Generator, frame level and sequence level Discriminator respectively. The idea of disentangled representation was explored by Zhou et al.\cite{zhou2019talking}. The authors proposed the Disentangled Audio-Visual System (DAVS), which uses disentangled audio-visual representation to create high-quality talking face videos.

\subsubsection{Video prediction}
Video prediction is the ability to predict future video frames based on the context of a sequence of previous frames. Formally future frame prediction can be defined as follows. Let $\mathbf{X}_{i} \in \mathrm{R}^{w \times h \times c}$ be the $i^{th}$ frame in the sequence of $n$ video frames $\mathbf{X}=\left(X_{i-n}, \ldots, X_{i-1}, X_{i}\right)$, where $w$, $h$, and $c$ denote the width, height and the number of channels respectively. The goal is to predict the next sequence of frames $\mathbf{Y}=\left(\hat{Y}_{i+1}, \hat{Y}_{i+2}, \ldots, \hat{Y}_{i+m}\right)$ using the input $\mathbf{X}$.

Video prediction is a challenging task due to the complex task of modelling both the content and motion in videos. To this extent several studies have been carried out to perform video prediction using GAN-based training (\cite{mathieu2016deep},\cite{villegas2018decomposing},\cite{aigner2018futuregan},\cite{VPGAN9022246},\cite{DualMotionGAN8237456},\cite{Vondrick8099802},\cite{FSTN8099713}).
Mathieu et al.\cite{mathieu2016deep} used multi-scale architecture for future frame prediction. The network was trained using an adversarial training method, and an image gradient difference loss function.
MCNet\cite{villegas2018decomposing} performs the task of video frame prediction by disentangling temporal and spatial dynamics in videos. An Encoder-Decoder Convolutional Neural Network is used to model video content and Convolutional-LSTM is used to model temporal dynamics or motion in the video. In this way predicting the next frame is as simple as converting the extracted content features into the next frame content using the recognized motion features.

FutureGAN\cite{aigner2018futuregan} used an encoder-decoder based GAN model to predict future frames of the video sequence. Their network comprises of Spatio-temporal 3D convolution network(3D ConvNets)\cite{7410867} for all encoder and decoder modules to capture both the spatial and temporal components of a video sequence. To have stable training and prevent problems of mode collapse the authors used Wasserstein GAN with gradient penalty (WGAN-GP)\cite{gulrajani2017improved} loss and the technique of progressively growing GAN or ProGAN\cite{PROGAN:DBLP:journals/corr/abs-1710-10196} which has been shown to generate high resolution images. VPGAN\cite{VPGAN9022246} is a GAN-based framework for stochastic video prediction. The authors introduce a new adversarial inference model, an action control conformal mapping network and use cycle consistency loss for their model. The authors also combined image segmentation models with their GAN framework for robust and accurate frame prediction. Their model outperformed other existing stochastic video prediction methods.  
With a unified architecture, Dual Motion GAN\cite{DualMotionGAN8237456} attempts to jointly resolve the future-frame and future-flow prediction problems. The proposed framework takes in as input a sequence of frames to predict the next frame by combining the future frame prediction with the future flow-based prediction. To achieve this, Dual Motion GAN employs a probabilistic motion encoder (to map frames to latent space), two Generators (a future-frame Generator and a future-flow Generator), as well as a flow estimator and flow-warping layer. A frame Discriminator and a flow Discriminator are used to classifying fake and real future frames and flow. Bhattacharjee et al.\cite{BhattacharjeeNIPS2017_b166b57d} tackle the problem of future frames prediction by using multi-stage GANs. To capture the temporal dimension and handle inter-frame relationships, the authors proposed two new objective functions, the Normalized Cross-Correlation Loss (NCCL) and the Pairwise Contrastive Divergence Loss
(PCDL). The multistage GAN(MS-GAN) is made up of 2 GANs to generate frames at two separate resolution whereby Stage-1 GAN output is fed to Stage-2 GAN to produce the final output. Kwon et al.\cite{8953708} proposed a novel framework based on CycleGAN\cite{CycleGAN2017} called the Retrospective CycleGAN to predict future frames that are farther in time but are relatively sharper than frames generated by other methods. The framework consists of a single Generator and two Discriminators. During training, the Generator is tasked with generating both future and past frames, and the retrospective cycle constraints are used to ensure bi-directional prediction consistency. The frame Discriminator detects individual fake frames, whereas the sequence Discriminator determines whether or not a sequence contains fake frames. According to the authors, the sequence Discriminator plays a crucial role in predicting temporally consistent future frames.  To train their model, the combination of two adversarial and two reconstruction loss functions were used. 

\begin{table}[h!]
\caption{Applications in Video generation and prediction}
\begin{tabular}{p{0.5\textwidth}p{0.45\textwidth}}
\hline
\textbf{Video generation and prediction} & \textbf{GAN models}       \\ \hline
Unconditional video generation        & VGAN \cite{VGAN10.5555/3157096.3157165}, TGAN \cite{TGAN}, FTGAN \cite{FTGAN:DBLP:journals/corr/abs-1711-09618}, MoCoGAN \cite{Tulyakov:2018:MoCoGAN}, DVD-GAN \cite{clark2019adversarial}             \\
Conditional video generation         &  VAE-GAN \cite{li2017video}, TGANs-C \cite{TGAN}, TFGAN \cite{TFGANijcai2019-276}, StoryGAN \cite{StoryGAN8953914}, BoGAN\cite{BoGAN9138828}, TiVGAN\cite{TiVGAN9171240}, LSTM and cGAN\cite{jalalifar2018speechdriven}, RNN based GAN\cite{Vougioukas_2019_CVPR_Workshops}, TemporalGAN\cite{zhou2019talking}  \\
Video Prediction                        & Multi-Scale GAN\cite{mathieu2016deep}, MCNet\cite{villegas2018decomposing}, FutureGAN\cite{aigner2018futuregan}, VPGAN\cite{VPGAN9022246}, Dual Motion GAN\cite{DualMotionGAN8237456}, MSGAN\cite{BhattacharjeeNIPS2017_b166b57d}, Retrospective Cycle GAN\cite{8953708}  \\ 
\hline
\end{tabular}
\label{tab:Video }
\end{table}

\subsection{Medical and Healthcare}
GANs have immense medical image generation applications and can be used to improve early diagnosis, reduce time and expenditure. Because medical image data is generally limited, GANs can be employed as data augmentation techniques by conducting image-to-image translation, synthetic data synthesis, and medical image super-resolution. 

One major application of GANs in medical and healthcare is the 
\textbf{image-to-image translation} framework, that is when multi-modal images are required one can use images from one modality or domain to generate images in another domain. Magnetic resonance imaging (MRI), is considered the gold standard in medical imaging. Unfortunately, it is not a viable option for patients with metal implants, as the metal in the machine could interfere with the results and the patients’ safety. MR-GAN\cite{s19102361} is similar to CycleGAN\cite{CycleGAN2017} and is used to transform 2D brain CT image slices into 2D brain MR image slices. However, unlike CycleGAN, which is used for unpaired image-to-image translation, MR-GAN is trained using both paired and unpaired data and combines adversarial loss, dual cycle-consistent loss, and voxel-wise loss. MCML-GANs\cite{Yu2019} uses the approach of multiple-channels-multiple-landmarks (MCML) input to synthesize color Retinal fundus images from a combination of vessel tree, optic disc, and optic cup images. The authors used two models based on the pix2pix\cite{Pix2Pix8100115} and CycleGAN\cite{CycleGAN2017} model framework and proposed several different architectures for the Generator model and compared their performance. Zhao et al.\cite{ZHAO201814} also proposed Tub-GAN and Tub-sGAN image-to-image translation framework to generate retinal and neuronal images. Armanious et al.\cite{ARMANIOUS2020101684} proposed the MedGAN framework for generalizing image to image translation in the field of medical image generation by combining the adversarial framework with a new combination of non-adversarial losses along with the usage of CasNet a ResNets\cite{DBLP:journals/corr/HeZRS15} inspired architecture. Sandfort et al.\cite{Sandfort2019} used CycleGAN\cite{CycleGAN2017} to transform contrast CT images into noncontrast images. The authors compared the segmentation performance of a U-Net trained on the original dataset versus a U-Net trained on a combined dataset of original data and synthetic non-contrast images were compared.

DermGAN\cite{48700} is used to generate synthetic images with skin conditions. The model learns to convert a semantic map containing a pre-specified skin condition, its size and location, as well as the underlying skin colour, into a realistic image that retains the pre-specified traits. The DermGAN Generator uses a modified U-Net\cite{U-NET10.1007/978-3-319-24574-4_28} where the deconvolution layers are replaced with a nearest-neighbor resizing layer followed by a convolution layer to reduce the checkerboard effect. The Generator and Discriminator are trained to minimize the combination of feature matching loss,  min-max GAN loss, $l_1$ reconstruction loss for the whole image, $l_1$ reconstruction loss for the pathological region. 

Apart from solving the image to image translation problems GAN are widely used for \textbf{synthetic medical image generation}(\cite{Middel10.1007/978-3-030-32689-0_13},\cite{8363564},\cite{8055572},\cite{8419363}).  Costa et al.\cite{8055572} implemented an adversarial autoencoder for the task of conditional retinal vessel network synthesis. Beers et al.\cite{DBLP:journals/corr/abs-1805-03144} applied ProGAN to generate high resolution and high quality  512x512 retinal fundus images and 256x256 multimodal glioma images. Zhang et al.\cite{8419363} used DCGAN\cite{radford2016unsupervised}, WGAN\cite{WGAN_pmlr-v70-arjovsky17a} and boundary equilibrium GANs (BEGANs)\cite{berthelot2017began} to generate synthetic medical images. They used the generated synthetic images to augment their datasets to build models with higher tissue recognition accuracy. Overall training with augmented datasets saw an increase in tissue recognition accuracy for the three GAN models when compared to baseline models trained without data augmentation. fNIRS-GANs\cite{Nagasawa_2020} based on WGAN\cite{WGAN_pmlr-v70-arjovsky17a} is used to perform  functional near-infrared spectroscopy (fNIRS) data augmentation to improve the fNIRS-brain–
computer interface (BCI) accuracy. Using data augmentation the authors were able to achieve higher classification accuracy of 0.733 and 0.746 for both the SVM and neural network models respectively as compared to 0.4 for both the models trained without data augmentation. To prevent data leakage by generating anonymized synthetic electrocardiograms (ECGs), Piacentino et al.\cite{Piacentinoelectronics10040389} used GANs. The authors first propose a new general procedure to convert raw data into images, which are well suited for GANs. Following that, a GAN design was established, trained, and evaluated. Because of its simplicity, the authors chose to use Auxiliary Classifier Generative Adversarial Network(ACGAN)\cite{ACGANpmlr-v70-odena17a} for their intended task. Kwon et al.\cite{2021PLoSO..1650458K} used GANs to augment mRNA samples to improve classification accuracy of deep learning models for cancer detection. With 5 fold increase in training data by combining GAN generated synthetic samples with the original dataset, the authors were able to improve the F1 score by 39\%.

\textbf{Image reconstruction and super resolution} are vital to obtaining high resolution images for diagnosis as due to constraints such as the amount of radiation used for MRI and other image acquisition techniques can highly impact the quality of images obtained. 
Multi-level densely connected super-resolution network, mDCSRN-GAN\cite{Chen2018EfficientAA} proposed by Chen et al. uses an efficient 3D neural network design for the Generator architecture to perform image super resolution.
MedSRGAN\cite{Gu2020} is an image super resolution (SR) framework for medical images. The authors used a novel convolutional neural network, Residual Whole Map Attention Network
(RWMAN) as the Generator network to low resolution features and then performs upsampling. For the Discriminator instead of having just the single generated high resolution image the authors used pairs of input low resolution images and generated high resolution images.  
Yamashita et. al\cite{10.1007/978-3-030-50426-7_37} evaluated several super resolution GAN models(SRCNN\cite{SRCNN_7115171}, VDSR\cite{VDSR_7780551}, DRCN\cite{DRCN_7780550} and ESRGAN\cite{ESRGAN10.1007/978-3-030-11021-5_5}) for Optical Coherence Tomography (OCT)\cite{OCTFujimoto2000} image enhancement. The authors found ESRGAN has the worst performance in terms of PSRN and SSIM but qualitatively, it was the best one producing sharper and high contrast images.

\begin{table}[h!]
\caption{Applications in Medical and Healthcare}
\begin{tabular}{p{0.5\textwidth}p{0.45\textwidth}}
\hline
\textbf{Medical and Healthcare application} & \textbf{GAN models}       \\ \hline
Multimodal Image to image translation         & pix2pix\cite{Pix2Pix8100115}, CycleGAN\cite{CycleGAN2017}, MR-GAN\cite{s19102361},MCML-GANs\cite{Yu2019}, Tub-GAN and Tub-sGAN\cite{ZHAO201814}, MedGAN \cite{ARMANIOUS2020101684}, DermGAN\cite{48700} \\
Image generation for data augmentation        & 
WGAN\cite{WGAN_pmlr-v70-arjovsky17a},
WGAN-GP\cite{NIPS2017_892c3b1c},
DCGAN \cite{radford2016unsupervised},
BEGAN\cite{berthelot2017began}, ProGAN \cite{PROGAN:DBLP:journals/corr/abs-1710-10196},
fNIRS-GANs\cite{Nagasawa_2020}, ACGAN\cite{ACGANpmlr-v70-odena17a}
\\
Image reconstruction                          &  SRCNN\cite{SRCNN_7115171}, VDSR\cite{VDSR_7780551}, DRCN\cite{DRCN_7780550}, ESRGAN\cite{ESRGAN10.1007/978-3-030-11021-5_5} mDCSRN-GAN \cite{Chen2018EfficientAA},MedSRGAN \cite{Gu2020}                          \\ 
\hline
\end{tabular}
\label{tab:Medical }
\end{table}

\subsection{Biology}
Biology is an area where generative models especially GANs can have a great impact by performing tasks such as protein sequence design, data augmentation and imputation and biological image generation. Apart from this GANs can also be applied for binding affinity prediction.

\textbf{Protein engineering} the process of identifying or developing useful or valuable proteins sequences with certain optimized properties. Several works have been done in relation to the application of Deep Generative models for protein sequence, especially the use of GANs(Repecka et al.\cite{Repecka2021}, Amimeur et al.\cite{Amimeur2020.04.12.024844} and Gupta et al.\cite{Gupta2019} ). GANs can be used to generate novel valid functional protein sequences and optimize protein sequences to have certain specific properties. ProteinGAN\cite{Repecka2021} can learn to generate diverse functional protein sequences directly from complex multidimensional amino acids sequence space. The authors specifically used GANs to generate functional malate dehydrogenases. 
Amimeur et al.\cite{Amimeur2020.04.12.024844} developed the Antibody-GAN which uses a modified WGAN for the generation of both single-chain and paired-chain antibody sequence generation. Their model is capable of generating extremely large diverse libraries of novel libraries that mimic somatically hypermutated human repertoire response. The authors also demonstrated the use of transfer learning to use their GAN model to generate molecules with specific properties of interest like MHC class II binding and specific complementarity-determining region (CDR) characteristics. FBGAN\cite{Gupta2019} uses the WGAN architecture along with the analyzer in a feedback-loop mechanism to optimize the synthetic gene sequences for desired properties using an external function analyser. The analyzer is a differential neural network and assigns a score to sampled sequences from the Generator. As training progresses lowest scoring generated sequences are replaced by high scoring generated sequences for the entire Discriminator's training set. GANs were utilised by Anand at al.\cite{NEURIPS2018_afa299a4} to generate protein structures, with the goal of using them in quick \emph{de novo} protein design.

GANs have been used for \textbf{data augmentation and data imputation} in biology due to the lack of available biosamples or the cost of collecting such samples. Some recent works include the generation and analysis of single-cell RNA-seq(\cite{Ghahramani262501}, \cite{Marouf2020}). The authors of cscGAN\cite{Marouf2020} or conditional single-cell generative adversarial neural networks used GANs for the generation of realistic single-cell RNA-seq data. Wang et al.\cite{10.1093/bioinformatics/bty563} proposed GGAN, a GAN framework to impute the expression values of the unmeasured genes. To do this they used a conditional GAN to leverage the correlations between the set of landmark and target genes in expression data. The Generator takes the landmark gene expression as input and outputs the target gene expression. This approach leverages correlations between the set of landmark and target genes in expression data from projects like 1000 Genomes. Park et al.\cite{Park2020RNA-seq} applied GANs to predict the molecular progress of Alzheimer’s disease (AD) by successfully analyzing RNA-seq data from a 5xFAD mouse model of AD. Specifically, the authors successfully applied WGAN+GP\cite{gulrajani2017improved} to bulk RNA-seq data with fewer variations in gene expression levels and a smaller number of genes.
scIGAN\cite{scIGANs10.1093/nar/gkaa506} is a GAN-based framework for scRNA-seq imputation. scIGANs can use complex, multi-cell type samples to learn non-linear gene-gene correlations and train a generative model to generate realistic expression profiles of defined cell types.

GANs can also be used to \textbf{generate biological imaging data}.
CytoGAN\cite{CytoGANGoldsborough227645} or Generative Modeling of Cell Images, the authors evaluated the use of several GAN models such as DCGAN\cite{radford2016unsupervised}, LSGAN\cite{LSGAN8237566} and WGAN\cite{WGAN_pmlr-v70-arjovsky17a} for cell microscopy imaging, in particular morphological profiling. Through their experiments, they discovered that LSGAN was the most stable, resulting in higher-quality images than both DCGAN and WGAN.
GANs have also been used for the generation of realistic looking  electron microscope images (Han et al.\cite{8354184}) and the generation of cells
imaged by fluorescence microscopy(Oskin et al.\cite{osokin:hal-01611692}).

Predicting \textbf{binding affinities} is an important task in drug discovery though it still remains a challenge.
To aid in drug discovery by predicting binding affinity between drug and target Zhao et l.\cite{GANsDTA10.3389/fgene.2019.01243} devised the use of a semi-supervised GANs-based method. The researchers utilised two GANs to learn representations from raw protein and drug sequence data, and a convolutional regression network to predict affinity.

\begin{table}[h!]
\caption{Applications in Biology}
\begin{tabular}{p{0.5\textwidth}p{0.45\textwidth}}
\hline
\textbf{Applications in Biology} & \textbf{GAN models}       \\ \hline
Protein Engineering        &  ProteinGAN \cite{Repecka2021}, Antibody-GAN \cite{Amimeur2020.04.12.024844}, FBGAN \cite{Gupta2019}, DCGANs \cite{NEURIPS2018_afa299a4}         \\
Data augmentation and data imputation         & cscGAN \cite{Marouf2020},GGAN \cite{10.1093/bioinformatics/bty563}, scIGAN \cite{scIGANs10.1093/nar/gkaa506}  \\
GANs for Biological Image Synthesis
& CytoGAN \cite{CytoGANGoldsborough227645}, SGAN \cite{8354184}, DCGAN+Wasserstein loss \cite{osokin:hal-01611692}\\
Binding affinity prediction & GANsDTA \cite{GANsDTA10.3389/fgene.2019.01243} \\
\hline
\end{tabular}
\label{tab:Biology }
\end{table}

\subsection{Astronomy}
With the advent of Big-data, the amount of data publicly available to scientists for data-driven analysis is mind boggling. Every day terabytes of data are being generated by hundreds if not thousands of satellites across the globe. With powerful computing resources GANs have found their way into astronomy as well for tasks such as image translation, data augmentation and spectral denoising. 

The authors of RadioGAN\cite{RadioGAN10.1093/mnras/stz1534} based their GAN on the Pix2Pix model to perform \textbf{image to image translation} between two different radio survey datasets to recover extended flux densities. Their model recovers extended flux density for nearly half of the sources within a 20\% margin of error and learns more complex relationships between sources in the two surveys than simply convolving them with a different synthesised beam. Several other authors have also used image to image translation models such as Pix2Pix, Pix2PixHD to generate solar images(Dash et al., Park et al.\cite{Park_2019}, Kim et al.\cite{Kim2019}, Jeong et al.\cite{Jeong_2020}, Shin et al.\cite{Shin_2020}  etc.).

Apart from image-to-image related tasks, GANs have been extensively used to \textbf{generate synthetic data} in the astronomy domain. Smith et al.\cite{10.1093/mnras/stz2886} proposed SGAN to produce realistic synthetic eXtreme Deep Field(XDF) images similar to the ones taken by the Hubble Space Telescope. Their SGAN model has a similar architecture to DCGAN and can be used to generate synthetic images in astrophysics and other domain. Ullmo et al.\cite{ullmo2020encoding} used GANs to generate cosmological images to bypass simulations which generally require lots of computing resources and are quite expensive. Dia et al.\cite{dia2019galaxy} showed that GANs can replace expensive model-driven approaches to generate astronomical images. In particular, they used ProGANs along with Wasserstein cost function to generate realistic images of galaxies. 
ExoGAN\cite{Zingales_2018} which is based on the DCGAN framework\cite{radford2016unsupervised} is the first deep-learning approach
for solving the inverse retrieval of exoplanetary atmospheres. According to the authors, ExoGAN was found to be up to 300 times faster than a standard retrieval for large spectral ranges. ExoGAN is designed to work with a wide range of instruments and wavelength ranges without requiring any additional training. 
Fussell et al.\cite{Fussell10.1093/mnras/stz602} explored the use of DCGAN\cite{radford2016unsupervised} and StackGAN\cite{StackGAN8237891} in a chained fashion for generation of high-resolution synthetic galaxies images.

The authors of Spectra-GAN\cite{9109312} designed their algorithm for \textbf{spectral denoising.} Their algorithm is based on CycleGAN i.e. it has two Generators and two Discriminators, with the exception that instead of unpaired samples SpectraGAN used paired examples. The model comprises of  three loss functions: adversarial loss, cycle-consistent loss, and
generation-consistent loss.

\begin{table}[h!]
\caption{Applications in Astronomy }
\begin{tabular}{p{0.5\textwidth}p{0.45\textwidth}}
\hline
\textbf{Applications in Astronomy} & \textbf{GAN models}       \\ \hline
Image to image translation & RadioGAN\cite{RadioGAN10.1093/mnras/stz1534},pix2pix\cite{Pix2Pix8100115},pix2pixHD\cite{wang2018pix2pixHD}  \\ 
Image data generation and augmentation &  SGAN\cite{10.1093/mnras/stz2886}, DCGAN\cite{radford2016unsupervised}, ProGAN\cite{PROGAN:DBLP:journals/corr/abs-1710-10196}, ExoGAN\cite{Zingales_2018}        \\
Image denoising         & Spectra-GAN\cite{9109312} \\
\hline
\end{tabular}
\label{tab:Astronomy}
\end{table}

\subsection{Remote Sensing}
Using GANs for remote sensing applications can be broadly divided into the following main categories:
\begin{itemize}
    \item \textbf{Data generation or augmentation:} Lin et al.\cite{8059820} proposed multiple-layer feature-matching generative adversarial networks (MARTA GANs) for remote sensing data augmentation. MARTA GAN is based on DCGAN\cite{radford2016unsupervised} however while DCGAN could produce images with a 64 × 64 resolution, MARTA GAN can produce 256×256 remote sensing images. To generate high-quality samples of remote images perceptual loss and feature-matching loss were used for model training. 
    Mohandoss et al.\cite{mohandoss2020generating} presented the MSG-ProGAN framework that uses ten bands of Sentinel-2 satellite imagery with varying resolutions to generate realistic multispectral imagery for data augmentation. To help with training stability the authors
    based their model on the MSGGAN\cite{karnewar2020msggan}, ProGAN\cite{PROGAN:DBLP:journals/corr/abs-1710-10196} models and
    used WGAN-GP\cite{gulrajani2017improved} loss function. Thus, MSG-ProGAN can generate multispectral 256 × 256 satellite images instead of RGB images.
    \item \textbf{Super Resolution:} HRPGAN\cite{Sun_2020} uses a PatchGAN inspired architecture to convert low resolution remote sensing images to high resolution images. The authors did not use batch normalization to preserve textures and sharp edges of ground objects in remote sensing images. Also, ReLU activations were replaced with SELU activations for overall lower training loss and stable training. In addition, the authors used a new loss function consisting of the traditional adversarial loss, perceptual reconstruction loss and regularization loss to train their model. D-SRGAN\cite{Demiray2021} converts low resolution Digital Elevation Models (DEMs) to high-resolution DEMs. D-SRGAN is based on the SRGAN\cite{ledig2017photorealistic} model. For training, D-SRGAN uses the combination of adversarial loss and content loss.
    \item \textbf{Pan-Sharpening:} Liu et al.\cite{Psgan8451049} proposed PSGAN for solving the task of image pan-sharpening and carried out several experiments using different image datasets and different Generator architectures. PSGAN is superior to many popular pan-sharpening approaches in terms of generating high-quality pan-sharpened images with fine spatial details and high-fidelity spectral information under both low-scale and full-scale image settings, according to their experiments. Furthermore, the authors discovered that two-stream architecture is usually preferable to stacking and that the batch normalisation layer and the self-attention module are undesirable in pan-sharpening. Pan-GAN\cite{Pan-GANMA2020110} uses one Generator and two Discriminators for performing pan sharpening. The Generator is based on the PNN\cite{7115171} architecture but the image scale in the Generator remains the same in different layers. The spectral and spatial Discriminators are similar in structure but have different inputs. The generated HRMS image or the interpolated LRMS image is fed into the spectral Discriminator. The original panchromatic image or the single channel image generated by the generated HRMS image after average pooling along the channel dimension are the inputs for the spatial Discriminator. 
    \item \textbf{Haze removal and Restoration:} Edge-sharpening cycle-consistent adversarial network (ES-CCGAN)\cite{rs12244162} is a GAN-based unsupervised remote sensing image dehazing method based on the CycleGAN\cite{CycleGAN2017}. The authors used the unpaired image-to-image translation techniques for performing image dehazing. ES-CCGAN includes two generator networks  and two discriminant networks. The Generators use DenseNet\cite{8099726} blocks instead of the ResNet\cite{he2015deep} block to generate dehazed remote-sensing images with plenty of texture information. An edge-sharpening loss was designed to restore clear edges in the images in addition to the adversarial loss, cycle-consistency loss and cyclic perceptual-consistency loss. Furthermore, to preserve contour information, a VGG16\cite{simonyan2015deep} network was re-trained using remote-sensing image data to evaluate the perceptual loss. 
    To tackle the problem of lack of availability of pairs of clear images and corresponding haze images to train the model, Sun et al.\cite{9263540} proposed a cascade method combining two GANs. A learning-to-haze GAN(UGAN) learns to haze remote sensing images using unpaired clear and haze image sets. The UGAN then guides the learning-to-dehaze GAN (PAGAN) to learn how to dehaze UGAN hazed images.
    Wang et al.\cite{ID-GAN8313133} developed the Image Despeckling Generative Adversarial Network (ID-GAN) to restore speckled Synthetic Aperture Radar (SAR) images. Their proposed method uses an encoder-decoder type architecture for the Generator which performs image despeckling by taking a noisy image as input. The Discriminator follows a standard layout with a sequence of convolution, batch normalization and ReLU layers, sigmoid function to distinguish between real and synthetic images. The authors used a refined loss function which is made up of pixel-to-pixel Euclidean loss, perceptual loss, and adversarial loss, all combined with appropriate weights. 
    \item \textbf{Cloud Removal:} Several authors have used GANs for the removal of clouds contamination from remote sensing images(\cite{Cloud-Gan8519033},\cite{LI2020373},\cite{pan2020cloud},\cite{rs13061079}). CLOUD-GAN\cite{Cloud-Gan8519033} can translate cloudy images into cloud-free visible range images. CLOUD-GAN functions similar to CycleGAN\cite{CycleGAN2017} having two Generators and two Discriminators. The authors use the LSGAN\cite{LSGAN8237566} training method as it has been shown to generate higher quality images with a much more stable learning process compared to regular GANs. For thin cloud removal in multi-spectral images, Li et al.\cite{LI2020373} proposed a novel semi-supervised method called CR-GAN-PM, which combines Generative Adversarial Networks and a physical model of cloud distortion. There are three networks in the CR-GAN-PM: an extraction network, a removal network, and a discriminative network. The GAN architecture is made up of the removal and discriminative networks. A combination of adversarial loss, reconstruction loss, correlation loss and optimization loss was used for training CR-GAN-PM.

\end{itemize}

\begin{table}[h!]
\caption{Applications in Remote Sensing }
\begin{tabular}{p{0.5\textwidth}p{0.45\textwidth}}
\hline
\textbf{Applications in Remote Sensing} & \textbf{GAN models}       \\ \hline
Data generation or augmentation & MARTA GAN\cite{8059820},MSG-ProGAN\cite{mohandoss2020generating}  \\ 
Super Resolution & HRPGAN\cite{Sun_2020},D-SRGAN\cite{Demiray2021}  \\ 
Pan-Sharpening &  PSGAN\cite{Psgan8451049},PAN-GAN\cite{Pan-GANMA2020110} \\
Haze removal and Restoration & ES-CCGAN\cite{rs12244162},ID-GAN\cite{ID-GAN8313133}, Sun et al.\cite{9263540}   \\
Cloud removal &  CLOUD-GAN\cite{Cloud-Gan8519033},CR-GAN-PM\cite{LI2020373} \\
\hline
\end{tabular}
\label{tab:Remote Sensing}
\end{table}

\subsection{Material Science}
GANs have a wide range of applications in material science. GANs can be used to handle a variety of material science challenges such as Micro and crystal structure generation and design, Designing of complex architectured materials, Inorganic materials design, Virtual microstructure design and Topological design of metaporous materials for sound absorption.

Singh et al.\cite{Singh2018PhysicsawareDG} developed physics aware GAN model for the synthesis of binary microstructure images. The authors used three models to accomplish the task. The first model is the WGAN-GP\cite{gulrajani2017improved}. The second approach replaces the usual Discriminator in a GAN with an invariance checker, which explicitly enforces known physical invariances. The third model combines the first two to recreate microstructures that adhere to both explicit physics invariances and implicit restrictions derived from image data. Yang et al.\cite{10.1115/1.4041371} proposed a GAN-based framework for microstructural materials design. A Bayesian optimization framework is used to obtain the microstructure with desired
material property by processing the GAN generated latent variables. CrystalGAN\cite{CrystalGANnouira2019crystalgan} is a novel GAN-based framework for generating chemically stable crystallographic structures with enhanced domain complexity. The CrystalGAN model consists of three main components, a First step GAN, a Feature transfer procedure and the Second step GAN synthesizes. The First step GAN  resembles the cross-domain GAN and generates pseudo-binary samples
where the domains are mixed. The Feature transfer technique brings greater order complexity to the data generated from the samples obtained in the preceding stage. Finally, the second step GAN synthesizes ternary stable chemical compounds while adhering to geometric limitations.  Kim et al.\cite{doi:10.1021/acscentsci.0c00426} proposed leveraging a coordinate-based crystal representation inspired by point clouds to generate crystal structures using generative adversarial network. Their Composition-Conditioned Crystal GAN can generate materials with the desired chemical composition by conditioning the network with a one-hot encoded composition vector. 
Designing complex architectured materials is challenging and is heavily influenced by the experienced designers’ prior knowledge. To tackle this issue, Mao et al.\cite{Mao2020DesigningCA} successfully used GANs for the design of complex architectured materials. Millions of randomly generated architectured materials classified into different crystallographic symmetries were used to train their model. Their proposed model generates complex architectured designs that require no prior knowledge and can be readily applied in a wide range of applications. 

Dan et al. proposed MatGAN\cite{MatGANDan2020} is the first GAN model for efficient sampling of inorganic materials design space by generating hypothetical inorganic materials. MatGAN, based on WGAN\cite{WGAN_pmlr-v70-arjovsky17a} can learn implicit chemical compositional rules from existing materials, allowing them to generate hypothetical yet chemically sound molecules. Another similar work was carried out by Hu et al.\cite{sym12111889} where they used WGAN\cite{WGAN_pmlr-v70-arjovsky17a} to generate hypothetical inorganic materials consistent with the atomic combination of the training materials. 

Lee et al.\cite{https://doi.org/10.1002/eng2.12274} employed DCGAN\cite{radford2016unsupervised}, CycleGAN\cite{CycleGAN2017} and Pix2Pix\cite{Pix2Pix8100115} to generate realistic virtual microstructural graph images. KL-divergence, a similarity metric that is considerably below 0.1, confirmed the similarity between the GAN-generated and ground truth images.

GANs were used by Zhang et al.\cite{ZHANG2021109855} to greatly accelerate and improve the topological design of metaporous materials for sound absorption. Finite Element Method (FEM) simulation image data were used to train the model. The quality of the GAN generated designs was confirmed by FEM simulation and experimental evaluation, demonstrating that GANs are capable of generating metaporous material designs with satisfactory broadband absorption performance.

\begin{table}[h!]
\caption{Applications in Material Science}
\begin{tabular}{p{0.5\textwidth}p{0.45\textwidth}}
\hline
\textbf{Applications in Material Science} & \textbf{GAN models}       \\ \hline
Micro and crystal structure generation and design &   Hybrid (WGAN-GP\cite{gulrajani2017improved}+GIN) \cite{Singh2018PhysicsawareDG}, GAN+GP-Hedge Bayesian optimization framework\cite{10.1115/1.4041371},  CrystalGAN\cite{CrystalGANnouira2019crystalgan}, Composition-Conditioned Crystal GAN\cite{doi:10.1021/acscentsci.0c00426}\\
Designing complex architectured materials & GAN-based model\cite{Mao2020DesigningCA}\\
Inorganic materials design & MatGAN\cite{MatGANDan2020}, WGAN\cite{WGAN_pmlr-v70-arjovsky17a} based model\cite{sym12111889}\\
Virtual microstructure design & (DCGAN\cite{radford2016unsupervised}, CycleGAN\cite{CycleGAN2017} and Pix2Pix\cite{Pix2Pix8100115})\cite{https://doi.org/10.1002/eng2.12274}  \\
Topological design of metaporous materials for sound absorption & GAN based model\cite{ZHANG2021109855} \\
\hline
\end{tabular}
\label{tab:Material Science}
\end{table}

\subsection{Finance}

Financial data modeling is a challenging problem as there are complex statistical properties and dynamic stochastic factors behind the process. Many financial data are time-series data, such as real property price and stock market index. Many of them are very expensive to available and usually do not have enough labeled historical data, which greatly limits the performance of deep neural networks. In addition, unlike the static features, such as gender and image data, time-series data has a high temporal correlation across time. This becomes more complicated when we model multivariate time series where we need to consider the potentially complex dynamics of these variables across time. Recently, with the development and wide usage of GAN in image and audio tasks, a lot of research works have proposed to generate realistic time-series synthetic data in finance.

Efimov et al.\cite{efimov2020using} combine conditional GAN (CGAN) and Deep Regret Analytic Generate Adversarial Networks (DRAGANs) to replicate three American Express datasets with high fidelity. A regularization term is added in the Discriminator loss in DRAGANs to avoid gradient exploding or gradient vanishing effects as well as to stabilize the convergence. 
Zhou et al.\cite{zhou2018stock} adopt the GAN-FD model (A GAN model for minimizing both forecast error loss and direction prediction loss) to predict stock prices. The Generator is based on LSTM layers while the Discriminator is using CNN layers. 
Li et al.\cite{li2020generating} propose a conditional Wasserstin GAN (WGAN) named Stock-GAN to capture history dependency for stock market order streams. The proposed Generator network has two crafted features (1) approximating the double auction mechanism underlying stock exchanges and (2) including the order-book features as the condition information. 
Wiese et al.\cite{wiese2020quant} introduce Quant GANs which use the Temporal Convolutional Networks (TCNs) architecture, also known as WaveNet\cite{oord2016wavenet} as the Generator. It shows the capability of capturing the long-range dependence such as the presence of volatility clusters in stock data such as S\&P 500 index.
FIN-GAN\cite{takahashi2019modeling} captures the temporal structures of financial time-series so as to generate the major stylized facts of price returns, including the linear unpredictability, the fat-tailed distribution, volatility clustering, the leverage effects, the coarse-fine volatility correlation, and the gain/loss asymmetry. 
Leangarun et al.\cite{leangarun2018stock} build LSTM-GANs to detect the abnormal trading behaviors caused by stock price manipulations. The base architecture for both Generator and Discriminator is LSTM. The simulated manipulation cases are used for testing purposes. The detection system was tested with the trading data from the Stock Exchange of Thailand (SET) which achieves 68.1\% accuracy in detecting pump-and-dump manipulations in unseen market data.

\begin{table}[h!]
\caption{Applications in Finance}
\begin{tabular}{p{0.5\textwidth}p{0.45\textwidth}}
\hline
\textbf{Applications in Finance} & \textbf{GAN models}       \\ \hline
Financial Data Generation &  CGAN and DRAGANs\cite{efimov2020using}, FIN-GAN\cite{takahashi2019modeling},
Stock-GAN\cite{li2020generating}\\ 
Stock Market Prediction &  GAN-FD\cite{zhou2018stock}, Quant GANs \cite{wiese2020quant}\\
Anomaly Detection in Finance &  LSTM-GANs\cite{leangarun2018stock}\\
\hline
\end{tabular}
\label{tab:Finance}
\end{table}

\subsection{Marketing} GANs can be leveraged to help businesses create effective marketing tools by synthesizing novel and unique designs for logos and generate fake images of models.

Typically designing a new logo is a fairly long and exhausting process and requires a lot of time and effort of the designer to meet the specific requirements of the clients. Sage et al.\cite{Sage8578714} put forward iWGAN, a GAN-based framework for virtually generating infinitely many variations of logos by specifying parameters such as shape, colour, and so on, in order to facilitate and expedite the logo design process. The authors proposed clustered GAN model to train on multi-modal data. Clustering was used to stabilise GAN training, prevent mode collapse and achieve higher quality samples on unlabeled datasets. The GAN models were based on the DCGAN\cite{radford2016unsupervised} and WGAN-GP\cite{gulrajani2017improved} models. LoGAN\cite{LoGAN8614182} or the  Auxiliary Classifier Wasserstein Generative Adversarial Neural Network with gradient penalty (AC-WGAN-GP) is based on the ACGAN\cite{ACGANpmlr-v70-odena17a} architecture. LoGAN can be used to generate logos conditioned on twelve predetermined colors. LoGAN consists of a Generator, a Discriminator and a classification network to help the Discriminator in classifying the logos. The authors use the WGAN-GP\cite{gulrajani2017improved} loss function for better training stability instead of using the ACGAN loss. 

GANs can be used to replace real images of people for marketing-related ads, by generating synthetic images and videos thus alleviate problems related to privacy. Ma et al.\cite{PG2NIPS2017_34ed066d} proposed Pose Guided Person Image Generation Network(PG$^2$) to generate synthetic fake images of a person in arbitrary poses conditioned on an image of a person and a new pose. PG$^2$ uses a two-stage process: Stage 1 Pose integration and Stage 2 Image refinement. Stage 1 generates a coarse output based on the input image and the target pose that depicts the human's overall structure. Stage 2 adopts the DCGAN\cite{radford2016unsupervised} model and refine the initial result through adversarial training, resulting in sharper images. 
Deformable GANs\cite{Siarohin2018DeformableGF} generate person images based on their appearance and pose. The authors introduced deformable skip connections and nearest neighbor loss to address large spatial deformation and to fix misalignment between the generated and ground-truth images. Song et al.\cite{Song_2019_CVPR} proposed E2E which uses GANs for unsupervised pose-guided person image generation. The authors break down the difficult task of learning a direct mapping under various poses into semantic parsing transformation and appearance generation to deal with its complexity.

\begin{table}[h!]
\caption{Applications in Marketing}
\begin{tabular}{p{0.5\textwidth}p{0.45\textwidth}}
\hline
\textbf{Applications in Marketing} & \textbf{GAN models}       \\ \hline
Logo generation & iWGAN\cite{Sage8578714},LoGAN\cite{LoGAN8614182}  \\ 
Model generation and pose generation & PG$^2$\cite{PG2NIPS2017_34ed066d}, Deformable GANs\cite{Siarohin2018DeformableGF}, E2E\cite{Song_2019_CVPR}  \\ 
\hline
\end{tabular}
\label{tab:Marketing}
\end{table}

\subsection{Fashion Designing} Fashion design is not the first thing that comes to mind when we think of GANs, however developing designs for clothes and outfits is another area where GANs have been utilized (\cite{10.1145/3311823.3311863},\cite{Attribute-GANLIU2019156},\cite{yuan2020garment}). 

Based on the cGAN\cite{cGANmirza2014conditional}, Attribute-GAN\cite{Attribute-GANLIU2019156} learns a mapping from a pair of outfits based on clothing attributes. The model has one Generator and two Discriminators. The Generator uses a U-Net architecture. A PatchGAN\cite{DBLP:journals/corr/IsolaZZE16} Discriminator is used to capture local high-frequency structural information and a multi-task attribute classifier Discriminator is used to determine whether the generated fake apparel image has the expected ground truth attributes. Yuan et al.\cite{yuan2020garment} presented Design-AttGAN, a new version of attribute GAN (AttGAN)\cite{8718508} to edit garment images automatically based on certain user-defined attributes. The AttGAN's original formulation is changed to avoid the inherent conflict between the attribute classification loss and the reconstruction loss.

\begin{table}[h!]
\caption{Applications in Fashion Design}
\begin{tabular}{p{0.5\textwidth}p{0.45\textwidth}}
\hline
\textbf{Applications in Fashion Design} & \textbf{GAN models}       \\ \hline
Clothes and garment design & P-GANs\cite{10.1145/3311823.3311863},Attribute-GAN\cite{Attribute-GANLIU2019156},Design-AttGAN\cite{yuan2020garment}  \\ 
\hline
\end{tabular}
\label{tab:Fashion Design}
\end{table}

\subsection{Sports} 
GANs can be used to generate sports texts, augment sports data, and predict and simulate sports activity to overcome the lack of labeled data and get insights into player behaviour patterns.

Li et al.\cite{10.1007/978-3-030-23407-2_9} used WGAN-GP\cite{gulrajani2017improved} for automatic generation of sport news based on game stats. Their WGAN model takes scores as input and outputs sentences describing how one team defeated another. This paper also showed the potential applications of GANs in the NLP area. 

JointsGAN\cite{9025417} was proposed by Li et al. to augment soccer videos with a dribbling actions dataset for improving the accuracy of the dribbling styles classification model. The authors use Mask R-CNN\cite{Mask-R-CNN8237584} and OpenPose\cite{8765346} to build dribbling player’s joint model to act as a condition and guide the GAN model. The accuracy of classification is improved from 88.14\% percent to 89.83\% percent using the dribbling player's joints model as the condition to the GAN. 
Theagarajan et al.\cite{Triplet-CNN-DCGAN} used GANs to augment their dataset to build robust deep learning classification, object detection and localization models for soccer-related tasks. Specifically, they proposed the Triplet CNN-DCGAN framework to add more variability to the training set in order to improve the generalisation capacity of the aforementioned models. The GAN-based model is made of a regularizer CNN (i.e., Triplet CNN) along with DCGAN\cite{radford2016unsupervised} and uses the DCGAN loss and the binary cross-entropy loss of the Triplet CNN. 

Memory augmented Semi-Supervised Generative Adversarial Network (MSS-GAN)\cite{MSS-GAN} can be used to predict the shot type and location in tennis. MSS-GAN is inspired by SS-GAN\cite{denton2016semisupervised}, coupled with memory modules to enhance its capabilities. The Perception Network (PN) is used to convert input images into embeddings, which are then combined with embeddings from the Episodic Memory (EM) and Semantic Memory (SM) to predict the next shot via the Response Generation Network (RGN). Finally, a GAN framework is used to train the network, with the RGN's predicted shot being passed to the Discriminator, which determines whether or not it is a realistic shot. BasketballGAN\cite{BasketballGAN} is a cGAN\cite{cGANmirza2014conditional} and WGAN\cite{WGAN_pmlr-v70-arjovsky17a} based framework to generate basketball set plays based on an input condition(offensive tactic sketched by coaches) and a random noise vector. The network was trained by minimizing the adversarial loss (Wasserstein loss\cite{WGAN_pmlr-v70-arjovsky17a}) dribbler loss, defender loss, ball passing loss, and acceleration loss.

\begin{table}[h!]
\caption{Applications in Sports}
\begin{tabular}{p{0.45\textwidth}p{0.5\textwidth}}
\hline
\textbf{Applications in Sports} & \textbf{GAN models}       \\ \hline
Sports text generation(NLP task) & WGAN-GP\cite{gulrajani2017improved} based GAN\cite{10.1007/978-3-030-23407-2_9} \\ 
Sports data augmentation &  JointsGAN\cite{9025417}, Triplet CNN-DCGAN\cite{Triplet-CNN-DCGAN} \\ 
Sports action prediction and simulation &  MSS-GAN\cite{MSS-GAN}, BasketballGAN\cite{BasketballGAN} \\
\hline
\end{tabular}
\label{tab:Sports}
\end{table}

\subsection{Music}
Because human perception is sensitive to both global structure and fine-scale waveform coherence, music or audio synthesis is an intrinsically tough deep learning problem. As a result, synthesising music requires the creation of coherent raw audio waveforms that preserve both global and local structures. GANs have been applied for tasks such as music genre fusion and music generation.

FusionGAN\cite{FusionGAN8215561} is a GAN based framework for unsupervised music genre fusion. The authors proposed the use of a multi-way GAN based model and utilized the Wasserstein distance measure as the objective function for stable training. 

MidiNet\cite{MidiNetYang2017MidiNetAC} is a CNN-GAN based
model  to generate music with
multiple MIDI channels. The model uses \emph{conditioner} CNN to model the temporal dependencies by using the previous bar to condition the generation of the present bar, providing a powerful alternative to RNNs. The model features a flexible design that allows it to generate many genres of music based on input and specifications. Dong et al.\cite{MuseGANdong2017musegan} proposed Multi-track Sequential Generative Adversarial Networks for
Symbolic Music Generation and Accompaniment or MuseGAN. Based on GANs, MuseGAN can be used for symbolic
multi-track music generation. Dong et al.\cite{Dong2018ConvolutionalGA} demonstrated a unique GAN-based model for producing binary-valued piano-rolls by employing binary neurons\cite{BinaryNeurons} as a refiner network in the Generator's output layer. When compared to existing approaches, the generated outputs of their model with deterministic binary neurons have fewer excessively fragmented notes. GANSYNTH\cite{GANSYNTHengel2018gansynth}, based on GANs can generate high-fidelity and locally-coherent
audio. The proposed model outperforms the state-of-the-art WaveNet\cite{WaveNet45774} model in generating high fidelity audio while also being much faster in sample generation.  GANs were employed by Tokui\cite{Tokui2020CanGO} to create realistic rhythm patterns in unknown styles, which do not belong to any of the well-known electronic dance music genres in training data. Their proposed Creative-GAN model, uses the Genre Ambiguity Loss to tackle the problem of originality. 
Li et al.\cite{INCO-GANmath9040387} presented an inception model-based conditional generative adversarial network approach (INCO-GAN), which allows for the automatic synthesis of complete variable-length music. Their proposed model comprises of four distinct components: a conditional vector Generator (CVG), an inception model-based conditional GAN (INCO-GAN), a time distribution layer and an inception model\cite{Inception_net_7780677}. Their analysis revealed that the proposed method's music is remarkably comparable to that created by human composers, with a cosine similarity of up to 0.987 between the frequency vectors. Muhamed et al.\cite{Transformer-GANsMuhamed2021SymbolicMG} presented the Transformer-GANs model, which combines GANs with Transformers
to generate long, high-quality coherent music sequences. A pretrained SpanBERT\cite{Joshi2020SpanBERTIP} is used as the Discriminator and  Transformer-XL\cite{Dai2019TransformerXLAL} as the Generator. To train on long sequences, the authors use the Gumbel-Softmax technique\cite{Kusner2016GANSFS} to obtain a differentiable approximation of the sampling process and to keep memory needs reasonable, a variant of the Truncated Backpropagation Through Time (TBPTT) algorithm\cite{Sutskever2014SequenceTS} was utilised for gradient propagation over long sequences.

\begin{table}[h!]
\centering
\begin{tabular}{p{0.45\textwidth}p{0.5\textwidth}}
\hline
\textbf{Applications in Music} & \textbf{GAN models}       \\ \hline
Music genre fusion & FusionGAN\cite{FusionGAN8215561}
\\
Music generation & MidiNet\cite{MidiNetYang2017MidiNetAC}, MuseGAN\cite{MuseGANdong2017musegan}, Binary Neurons based WGAN-GP\cite{Dong2018ConvolutionalGA}, GANSYNTH\cite{GANSYNTHengel2018gansynth}, Creative-GAN\cite{Tokui2020CanGO}, INCO-GAN\cite{INCO-GANmath9040387}, Transformer-GANs\cite{Transformer-GANsMuhamed2021SymbolicMG} \\
\hline
\end{tabular}
\caption{Applications in Music}
\label{tab:Music}
\end{table}

\section{Limitations of GANs and Future direction}
In this section, we'll go through some of the issues that GANs encounter, notably those related to training stability. We also discuss some of the prospective research areas in which GAN productivity could be enhanced.
\subsection{Limitations of GANs}
Generative adversarial networks (GANs) have gotten a lot of interest because of their capacity to use a lot of unlabeled data. While great progress has been achieved in alleviating some of the hurdles associated with developing and training novel GANs, there are still a few obstacles to overcome. We explain some of the most typical obstacles in training GANs, as well as some proposed strategies that attempt to mitigate such issues to some extent.
\begin{enumerate}
    \item \textbf{Mode Collapse:} In most cases, we want the GAN to generate a wide range of outputs. For example, while creating photos of human faces we'd like the Generator to generate varied-looking faces with different features for every random input to Generator. Mode collapse happens when the Generator can produce only a single type of output or a small set of outputs. This may occur as a result of the Generator's constant search for the one output that appears most convincing to the Discriminator in order to easily trick the Discriminator, and hence continues to generate that one type. 
    
    Several approaches have been proposed to alleviate the problem of mode collapse. Arjovsky et al. \cite{WGAN_pmlr-v70-arjovsky17a} found that Jensen-Shannon divergence is not ideal for measuring the distance of the distribution of the disjoint parts. As a result, they proposed the use of Wasserstein distance to calculate the distance between the produced and real data distributions. Metz et al.\cite{metz2017unrolled} proposed Unrolled Generative Adversarial Networks, which limit the risk of the Generator being over-optimized for a certain Discriminator, resulting in less mode collapse and increased stability.
    
    \item \textbf{Non-convergence:} Although GANs are capable of achieving Nash equilibrium\cite{GoodfellowNIPS2014}, arriving at this equilibrium is not straightforward. The training procedure necessitates maintaining balance and synchronisation between the Generator and Discriminator networks for optimal performance. Furthermore, only in the case of a convex function can gradient descent guarantee Nash equilibrium.
    
    Adding noise to Discriminator inputs and penalising Discriminator weights (\cite{arjovsky2017principled}, \cite{roth2017stabilizing}) are two techniques of regularisation that authors have sought to utilise to improve GAN convergence.
    
    \item \textbf{Vanishing Gradients:} Generator training can fail owing to vanishing gradients if the Discriminator is too good. A very accurate Discriminator produces gradients around zero, providing little feedback to the Generator and slowing or stopping learning. 
    
    Goodfellow et al.\cite{GoodfellowNIPS2014} proposed a tweak to minimax loss to prevent the vanishing gradients problem. Although this tweak to the loss alleviates the vanishing gradients problem, it does not totally fix the problem, resulting in more unstable and oscillating training. The Wasserstein loss\cite{WGAN_pmlr-v70-arjovsky17a} is another technique to avoid vanishing gradients because it is designed to prevent vanishing gradients even when the Discriminator is trained to optimality.
    
\end{enumerate}

\subsection{Future direction}
Even though GANs have some limits and training issues, we simply cannot overlook their enormous potential as generative models. The most important area for future research is to make advancements in theoretical aspects to address concerns like mode collapse, vanishing gradients, non-convergence, and model breakdown.  Changing learning objectives, regularising objectives, training procedures, tweaking hyperparameters, and other techniques have been proposed to overcome the aforementioned problems, as outlined in section 5.1. In most cases, however, accomplishing these objectives entails a trade-off between desired output and training stability. As a result, rather than addressing one training issue at a time, future research in this field should use a holistic approach in order to achieve a breakthrough in theory to overcome the challenges mentioned above.

Besides overcoming the above theoretical aspects during model training, Saxena et al.\cite{saxena2021generative} highlight some promising future research directions. (1) Keep high image quality without losing diversity. (2) Provide more theoretical analysis to better understand the tractable formulations during training and make training more stable and straightforward. (3) Improve the algorithm to make training efficient (4) Combine other techniques, such as online learning, game theory, etc with GAN.

\section{Conclusion}
In this paper, we presented state-of-the-art GAN models and their applications in a wide variety of domains. GANs' popularity stems from their ability to learn extremely nonlinear correlations between latent space and data space. As a result, the large amounts of unlabeled data that remain closed to supervised learning can be used. We discuss numerous aspects of GANs in this article, including theory, applications, and open research topics.
We believe that this study will assist academic and industry researchers from various disciplines in gaining a full grasp of GANs and their possible applications. As a result, they will be able to analyse the potential application of GANs for their specific tasks.

\bibliographystyle{ACM-Reference-Format}
\bibliography{sample-base}

\end{document}